\documentclass[runningheads,a4paper]{llncs}

\usepackage{amssymb}
\usepackage{graphicx}
\usepackage{siunitx}
\usepackage{booktabs}
\usepackage{xcolor}

\usepackage{url}
\urldef{\mailsc}\path|{a.bokhovkin, e.burnaev}@skoltech.ru|    
\newcommand{\keywords}[1]{\par\addvspace\baselineskip
\noindent\keywordname\enspace\ignorespaces#1}

\begin{document}

\mainmatter  

\title{Boundary Loss for Remote Sensing Imagery Semantic Segmentation}

\titlerunning{Boundary Loss for Remote Sensing Imagery Semantic Segmentation}

%
%
\author{Alexey Bokhovkin$^1$
\and Evgeny Burnaev$^2$\thanks{The work was supported by The Ministry of Education and Science of Russian Federation, grant No. 14.615.21.0004, grant code: RFMEFI61518X0004.}}
\authorrunning{Boundary Loss for Remote Sensing Imagery Semantic Segmentation}

\institute{Skoltech, Moscow, Russia\\
$^1$Aeronet, $^2$ADASE\\
\mailsc}

%
%

\toctitle{Boundary Loss for Remote Sensing Imagery Semantic Segmentation}
\tocauthor{A. Bokhovkin, E. Burnaev}
\maketitle

\begin{abstract}
In response to the growing importance of geospatial data, its analysis including semantic segmentation becomes an increasingly popular task in computer vision today. Convolutional neural networks are powerful visual models that yield hierarchies of features and practitioners widely use them to process remote sensing data. When performing remote sensing image segmentation, multiple instances of one class with precisely defined boundaries are often the case, and it is crucial to extract those boundaries accurately. The accuracy of segments boundaries delineation influences the quality of the whole segmented areas explicitly. However, widely-used segmentation loss functions such as BCE, IoU loss or Dice loss do not penalize misalignment of boundaries sufficiently. In this paper, we propose a novel loss function, namely a differentiable surrogate of a metric accounting accuracy of boundary detection. We can use the loss function with any neural network for binary segmentation. We performed validation of our loss function with various modifications of UNet on a synthetic dataset, as well as using real-world data (ISPRS Potsdam, INRIA AIL). Trained with the proposed loss function, models outperform baseline methods in terms of IoU score.
\end{abstract}

\section{Introduction}\label{INTRODUCTION}

Semantic segmentation of remote sensing images is a critical process in the workflow of object-based image analysis, which aim is to assign each pixel to a semantic label \cite{RSdamage2018,cch19}. It has applications in environmental monitoring, urban planning, forestry, agriculture, and other geospatial analysis. Although a general image segmentation problem is relatively well investigated, we consider a particular type of this problem related to buildings segmentation.

The common aspect of urban aerial imagery is its high resolution (from 0.05 to 1.0 m). Higher GSD brings a lot of small details and structures, but also increases intra-class variance and decreases inter-class differences. This applies particularly to the class of buildings, including an enormous number of shapes and patterns. After the stage of segmenting buildings, further analysis could be done, such as distinct building classification and height estimation, population prediction, economic forecasting, etc. For most of the tasks, it is crucial to separate all buildings and minimize the number of false positive/false negative instances. Even with high-resolution imagery, it is difficult to segment every object separately, especially in high-density urban areas, although we get more accurate information about the boundaries of buildings. Therefore it is necessary to construct a method that increases the attention of a neural network, used for semantic segmentation, to borders of closely located objects.

CNNs for semantic segmentation typically use such loss functions as cross-entropy, dice score and direct $IoU$ optimization while training the network, but they are not sensitive enough to some misalignment of boundaries. Even being deviated from the ground truth by 5-10 pixels, predicted boundary does not significantly contribute to the value of the loss functions, mentioned above, and scores of common pixel-wise metrics. To better account for boundary pixels we select $BF_1$ metric (see original work \cite{bf1} and its extension \cite{bf2}) to construct a differentiable surrogate and use it in training. The surrogate is not used alone for training, but as a weighted sum with $IoU$ loss (from direct $IoU$ optimization). We found that the impact of the boundary component of the loss function should be gradually increased during training, and so we proposed a policy for the weight update. The assumption why this works is that for the first epochs network learns to label instances as shapeless blobs with very uncertain boundaries, and after finding an instance we only need to adjust its borders because they define the whole segments explicitly. 

In experiments we also found that network trained with a combination of $IoU$ and boundary loss converges faster: we increase the confidence of the network in its predictions on the border as fast as for intra-segment pixels; with other losses, the mask near the borders is very blurred. The next point is that network better handles edge effects while baseline models tend to miss or distort masks on the edge of an input image. Besides, if it is highly important to separate neighboring buildings, methods of instance segmentation are often applied, which can require extra data preparation from raw masks and they are commonly multistage. However, binary segmentation with boundary loss is end-to-end. In experiments on the challenging datasets ISPRS Potsdam and INRIA AIL, we obtain $>93.8\%$ and $>74.3\%$ pixel-wise $IoU$ score respectively. There is also a comparison with another method for accurate delineation of curvilinear structures \cite{vgg}, which encounters difficulties in remote sensing tasks. Our results show a consistent increase in the performance for all models on both datasets in comparison to various loss functions. 

The remainder of the paper is organized as follows. Section~\ref{RELATED_WORK} briefly reviews related work. The proposed boundary loss, as well as the steps to construct its surrogate, are presented in Section~\ref{METHOD}. Experiments and results are discussed in Section~\ref{EXPERIMENTS}, followed by concluding remarks in Section~\ref{CONCLUSIONS}. 

\section{Related work}\label{RELATED_WORK}

The subject of our work intersects with two branches of research, which are Deep Neural Networks (DNNs) and direct optimization of performance measures or their differentiable versions also called surrogates. Besides, we should mention methods for accurate boundary delineation, as one of them (see \cite{vgg}) was compared to our approach.

\textbf{DNNs and semantic segmentation.} Semantic segmentation based on DNNs is a pixel-wise mapping to semantic labels. Since 2012 a lot of neural networks \cite{long} for solving this task were constructed. The success of AlexNet \cite{alexnet} in ImageNet challenge marked the beginning of neural networks application in computer vision tasks such as classification, object detection, and semantic segmentation. Initially, for the latter problem, they adapted CNNs, used for classification. Namely, they remove fully connected layers, and use initial layers as an encoder to downsample an image and derive features that have stronger semantic information but low spatial resolution. Another part of the network called decoder obtains a high-resolution mask of labels, which upsamples features into the resulting mask. 

For the last few years, DNNs have shown to be very effective for semantic segmentation task. There are several models now that are regarded as state-of-the-art. First of all, it is SegNet architecture \cite{Badrinarayanan2016SegNetAD} which fits online video segmentation very well. DeepLab network \cite{DeepLab} uses atrous convolutions and CRFs at postprocessing step to refine small details of the segmentation. The FRRN  \cite{Pohlen2017FullResolutionRN} is an example of a model with multi-scale processing technique; it is based on two separate streams of data to better handle the high-level semantic information and the low-level pixel information simultaneously. Finally, UNet architecture \cite{unet} is very popular nowadays for its enormous flexibility. Among aforementioned models UNet remains preferable because of the high efficiency of feature extraction and ability to apply various neural network architectures as backbones. UNet is an asymmetric FCNN (Fully Convolutional Neural Network) with skip connections between the downsampling and upsampling paths. UNet has shown its high performance in competitions and research projects, and we use it for all experiments in the paper.

\textbf{Performance measure optimization.} 
There already exist papers concerning direct optimization of metrics or their differentiable versions. In various applications such measures as $IoU$, $F_1$-score, $ROC$-area, $mAP$ are widely used. Constructing surrogates of these measures is the most tricky part because it can be very hard or even impossible to replace non-differentiable operations with differentiable ones and keep computational efficiency at the same time. For the task of segmentation, $IoU$ is usually used to measure the performance of any segmentation approach. As a result, there exists a lot of its surrogates, and the goal is to minimize the gap between the actual $IoU$ value and its differentiable approximation. In paper \cite{neuro} they proposed $NeuroIoU$ loss, which approximates naive $IoU$-loss with a neural network. Another approach \cite{lovasz} called Lovasz-Softmax loss, is based on the convex Lovasz extension of submodular losses. A critical property for this surrogate is that it effectively captures the absolute minimum of the original loss.

In this paper, we construct a differentiable version of the metric, presented in \cite{bf1}. The motivation is that boundaries of segments explicitly define them and their extraction is necessary for accurate building segmentation. Authors of \cite{bf1} proposed a novel boundary metric $BF_1$ that accounts for accuracy of contour extraction and overcomes the weaknesses of mainstream performance measures. For $BF_1$ we will construct a differentiable surrogate and show how it increases the accuracy of segmenting borders.

Below we provide a list of metrics and loss functions to be used for comparative analysis:
\begin{itemize}
\item Direct $IoU$ loss: $IoU = \frac{TP}{FP+TP+FN},$ $L_{IoU} = 1-IoU$,
\item $Dice$ loss: $F_1 = \frac{2TP}{2TP+FN+FP},$ $L_{F_1} = 1-F_1$, 
\item $SS$ loss: $SS = \lambda \frac{TN}{TN+FP} + (1-\lambda)\frac{TP}{TP+FN},$ $L_{SS} = 1-SS,$
where $\lambda$ is a weight to balance two components, \item $VGG$ loss: implemented in \cite{vgg}.
\end{itemize}

The terms $TP,$ $FP,$ $TN,$ $FN$ denote pixel sets of true positive, false positive, true negative, false negative classes on a predicted binary map. For all loss functions, notations of corresponding metrics or approaches, inducing them, are provided in a subscript. By $L_{BF_1, IoU}$ we denote a weighted sum  \[wL_{BF_1} + (1-w)L_{IoU}\] for an arbitrary weight $w\in[0,1]$. 

\textbf{Accurate boundary delineation.} One possible solution to extract borders accurately is to use CRF (Conditional Random Fields). It works as an extra layer above the output of the original neural network. CRF is applied to capture additional contextual information and to produce much more refined prediction \cite{crf}. Authors of \cite{nowozin} used CRF together with Bayesian decision theory and proposed a heuristic to maximize the value of $EIoEU$. Another approach, a combination of CRF and superpixels \cite{superpixels}, takes advantage of superpixel edge information and the constraint relationship among different pixels. Proposed algorithm of boundary optimization made it possible to improve $IoU$ score by 3\% on PASCAL VOC 2012 \cite{pascal-voc-2012} and Cityscapes \cite{cityscapes} datasets compared to the performance of plain FCN.

We compare an approach of \cite{vgg} to the one, proposed in our paper. Authors claim that pixel-wise losses alone such as binary cross-entropy are unsuitable for predicting curvilinear structures. For this problem, they developed a new loss term based on features extracted with VGG19 network \cite{vgg19}. Conceptually their approach is similar to ours because we add our surrogate to the weighted conventional loss function. In contrast to VGG19 features, using our surrogate, we manually extract boundaries which are features too and encourage a neural network to draw attention to them much stronger.

\section{Our Method}\label{METHOD}

In this work, we use deep neural network UNet as a base model, because it is one of the most reliable and universal models. As we already pointed out in the introduction, there are a lot of different types of encoders and decoders that we can use as backbones. Of course, this network can be extended to two-headed version (UNet with two decoders), where the first head predicts whole segments and the other predicts only their boundaries. However, this approach requires additional computational resources and data preparation compared to the original UNet. Our interest is to propose a method that would not increase implementation and computational complexities. Further, we first introduce $BF_1$ metric from \cite{bf1} and then describe how to construct its differentiable surrogate.

\subsection{Boundary metric ($BF_1$)}\label{sec:Boundary metric}

Let $S_{gt}^c$, $S_{pd}^c$ be the binary maps of class $c$ in the ground truth and predicted segmentation respectively, $B_{gt}^c$, $B_{pd}^c$ --- the boundaries for these binary maps. Then the precision and the recall for class $c$ are defined as follows:
\[
P^c = \frac{1}{|B^c_{pd}|}\sum_{x\in B^c_{pd}}[[d(x, B_{gt}^c)<\theta]],\,
R^c = \frac{1}{|B^c_{gt}|}\sum_{x\in B^c_{gt}}[[d(x, B_{pd}^c)<\theta]],
\]
where brackets $[[\cdot]]$ denote indicator function of a logical expression, $d(\cdot)$ is Euclidean distance measured in pixels, $\theta$ - some predefined threshold on a distance; in all experiments we set $\theta$ to $3$ or $5$. Here distance is calculated from a point to one of two sets (ground truth or predicted boundary) as the shortest distance from the point to the point of the boundary. The $BF_1^c$ measure for class $c$ is defined as:
\[
BF_1^c = \frac{2P^cR^c}{P^c+R^c}.
\]

\subsection{Surrogate construction ($L_{BF_1}$)}\label{sec:Surrogate construction}

To construct the differentiable version of the metric $BF_1$, we need to extract the boundary of any segment somehow. There are several ways to do this. Let us denote by $y_{pd}$ a binary map for an arbitrary class $c$ for some image $I$, predicted by a neural network, $y_{gt}$ a ground truth map for the same class and the same image. We expect values of the predicted map to be distributed in $[0,1]$, and values of the ground truth map to be in $\{0, 1\}$. The boundaries can be defined as
\begin{equation}\label{eq:gt_b}
y^b_{gt} = pool(1-y_{gt}, \theta_0) - (1-y_{gt}),\,
y^b_{pd} = pool(1-y_{pd}, \theta_0) - (1-y_{pd}).
\end{equation}

Here $pool(\cdot, \cdot)$ applies a pixel-wise max-pooling operation to the inverted predicted or ground truth binary map with a sliding window of size $\theta_0$ (hyperparameter $\theta_0$ must be as small as possible to extract vicious boundary; usually we set $\theta_0=3$). Value $(1-y_{gt, pd})$ corresponds to inversion of any pixel of the map. To compute Euclidean distances from pixels to boundaries a supporting map should be obtained, which is the map of the extended boundary:
\begin{equation}\label{eq:gt_ext}
y^{b,ext}_{gt} = pool(y^b_{gt}, \theta),\, y^{b,ext}_{pd} = pool(y^b_{pd}, \theta).
\end{equation}

The value of hyperparameter $\theta$ can be determined as not greater than the minimum distance between neighboring segments of the binary ground truth map. After that precision and recall can be computed as follows:

\begin{equation}
P^c = \frac{sum(y_{pd}^b \circ y^{b,ext}_{gt})}{sum(y_{pd}^b)},\,
R^c = \frac{sum(y_{gt}^b \circ y^{b,ext}_{pd})}{sum(y_{gt}^b)},
\end{equation}
where operation $\circ$ denotes pixel-wise multiplication of two binary maps and operation $sum(\cdot)$ --- pixel-wise summation of a binary map. Finally, the reconstructed metric and a corresponding loss function are defined as:
\begin{equation}
BF_1^c = \frac{2P^cR^c}{P^c+R^c},\,
L_{BF_1^c} = 1-BF_1^c.
\end{equation}

All operations above are differentiable in terms of either derivative or subderivative. In Fig.~\ref{fig:demo_loss} there is an example of how the proposed surrogate works. It is worth to mention that of course we can use convolutions with edge detection filters, such as Sobel, but in experiments, we found that this approach is not as effective as max-pooling. Nevertheless, both methods have their drawbacks. The disadvantages of both methods are that on the first step we extract boundary not narrower than $2$-$3$ pixels in width, but with Sobel filters, boundaries are even wider and more blurred. The main drawback of the used max-pooling operation is that gradients are passed only in tensor cells with a maximum value within a current sliding window. Sobel operators do not have this problem and take into account all pixels of the sliding window. 

\section{Experiments}\label{EXPERIMENTS}

\subsection{Example on synthetic dataset (AICD)}\label{sec:Example on synthetic dataset}

To demonstrate capabilities of the novel boundary loss and test the hypothesis that $L_{BF_1}$ can assist $L_{IoU}$ a synthetic dataset was used. It consists of $800$ images with one primitive segment in each image. 
We trained on this dataset without augmentations and postprocessing a not deep fully convolutional neural network with four conv layers in the encoder and four conv layers in the decoder to show that $L_{BF_1, IoU}$ outperforms other losses under simple conditions. 

\begin{figure}[ht!]
\centering
\begin{tabular}{ccccccc}
\includegraphics[width=21mm, height=21mm]{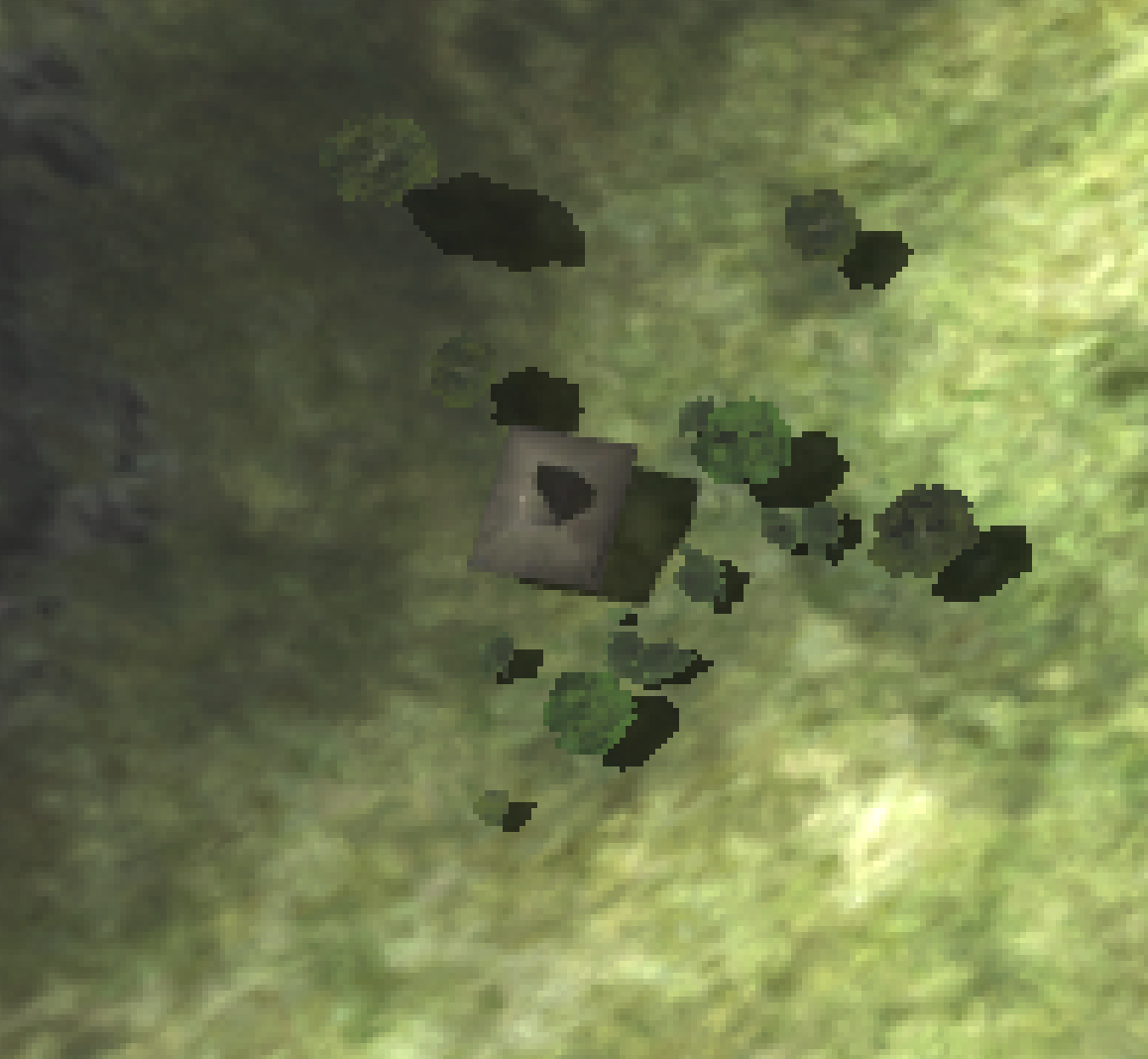} & \raisebox{0.33in}{\rotatebox[origin=t]{90}{\textit{Ground truth}}} & \includegraphics[width=21mm]{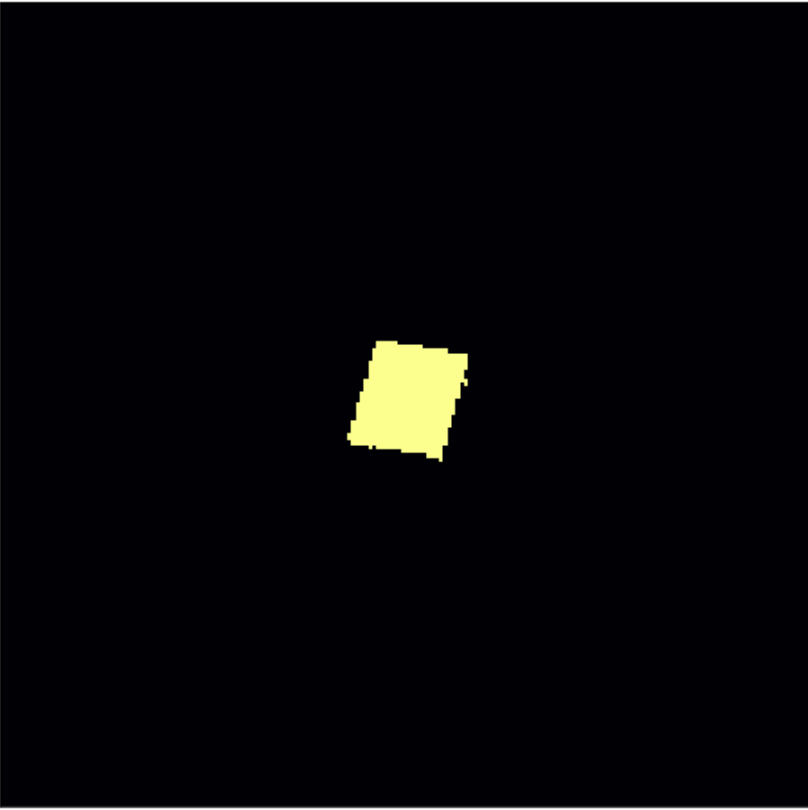} &   \includegraphics[width=21mm]{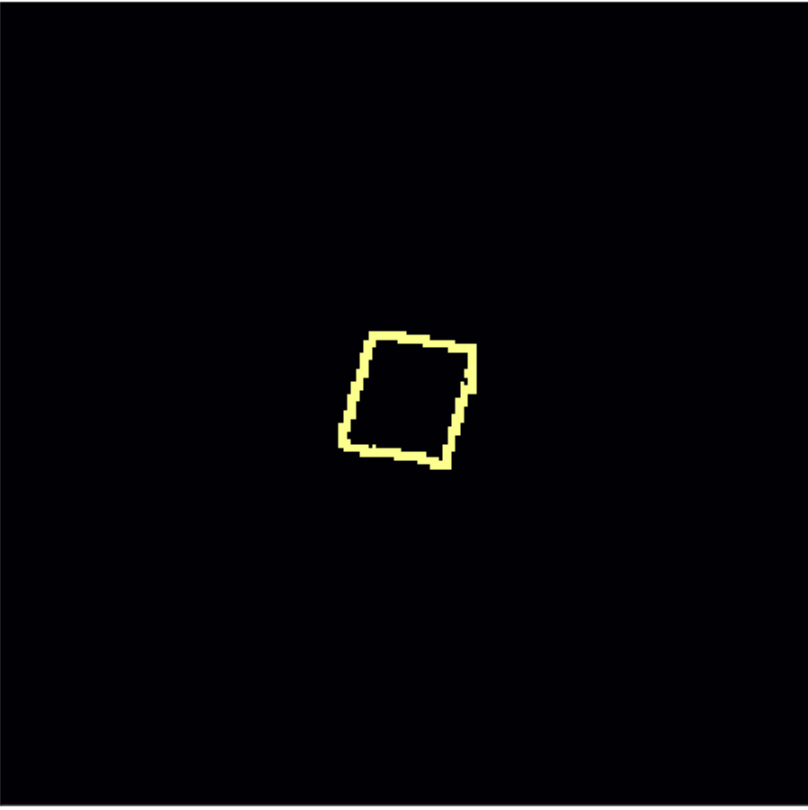} & \includegraphics[width=21mm]{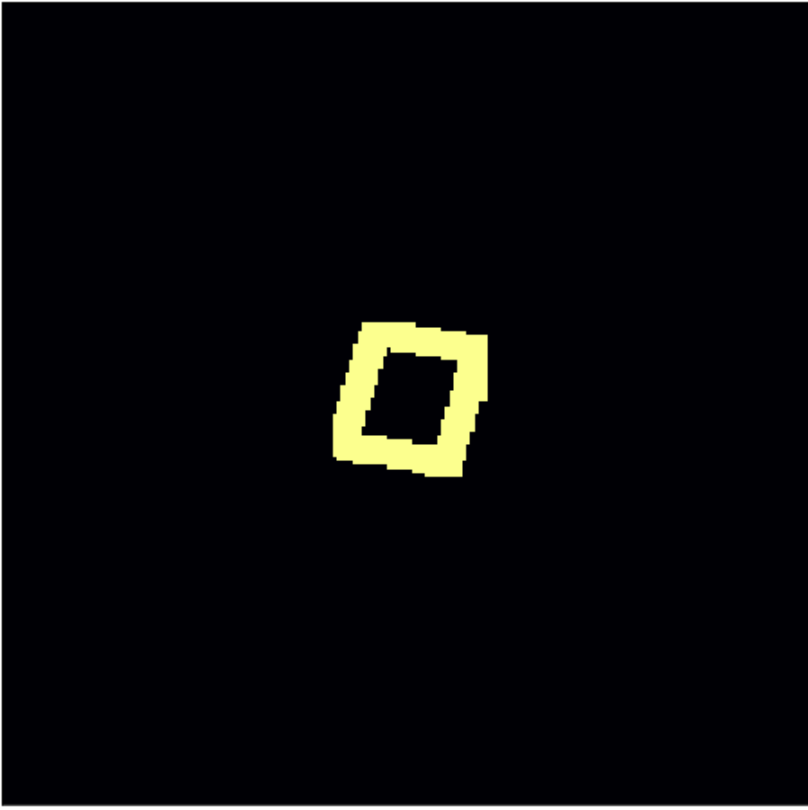} & \raisebox{0.33in}{\rotatebox[origin=t]{90}{\textit{Precision map}}} & \includegraphics[width=21mm]{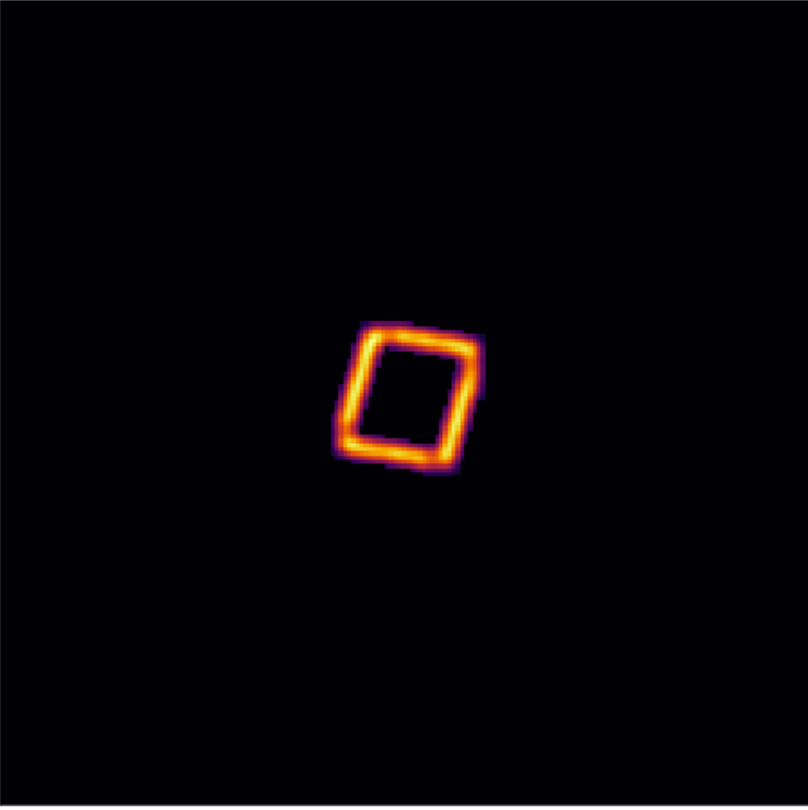}  \\
(a) & & (b) & (d) & (f) & & (h)  \\[6pt]
&
\raisebox{0.33in}{\rotatebox[origin=t]{90}{\textit{Prediction}}} & \includegraphics[width=21mm]{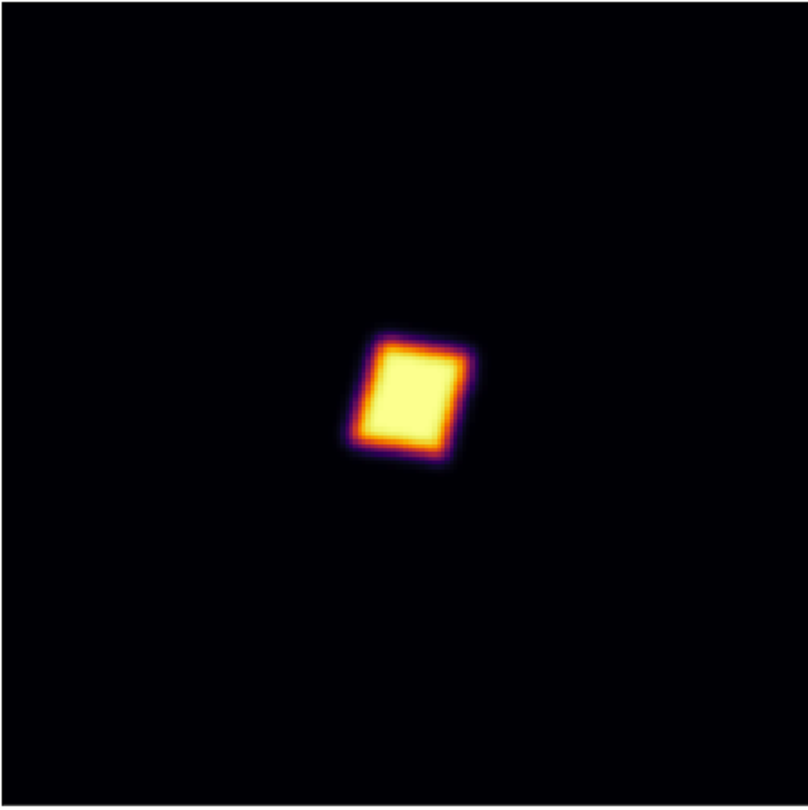} &   \includegraphics[width=21mm]{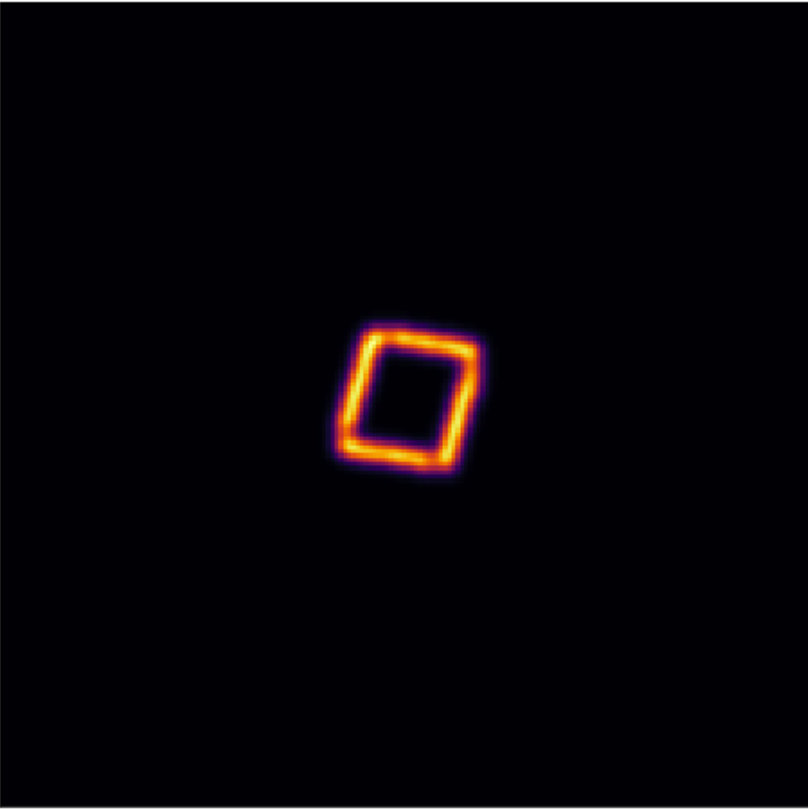} & \includegraphics[width=21mm]{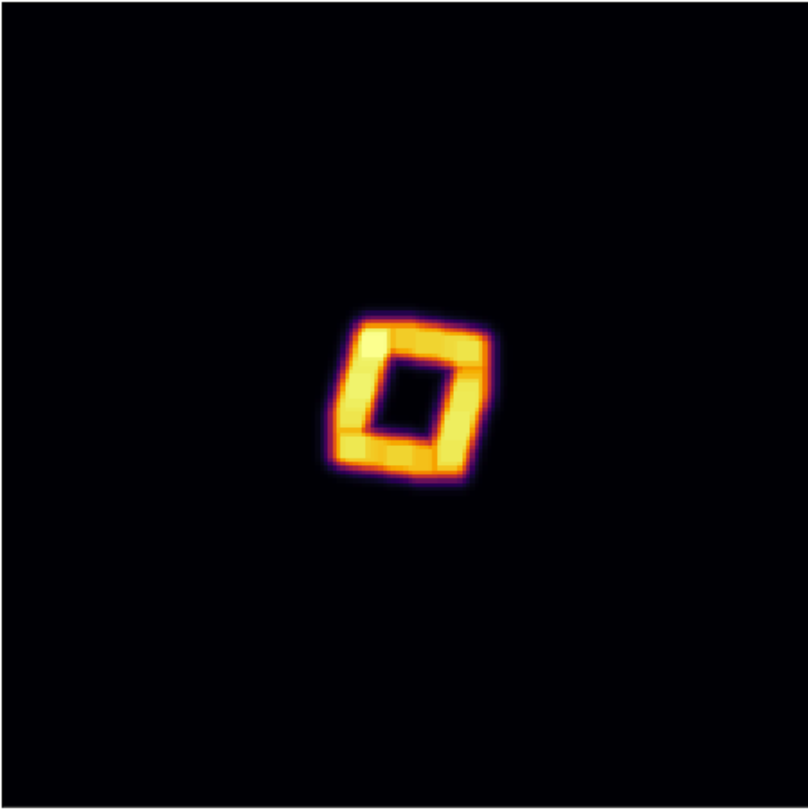} & \raisebox{0.33in}{\rotatebox[origin=t]{90}{\textit{Recall map}}} & \includegraphics[width=21mm]{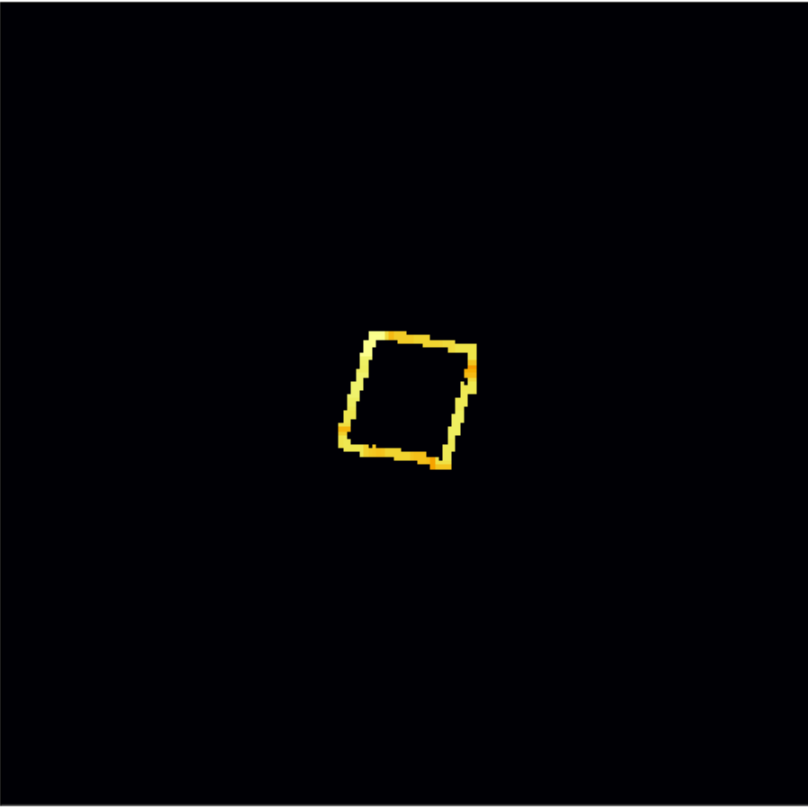}\\
& & (c) & (e) & (g) & & (i) \\[6pt]
\end{tabular}
\caption{(a) original orthophoto; (b) ground truth segment \texttt{(gt)}; (c) predicted segment \texttt{(pred)}; (d) boundary of \texttt{gt}; (e) boundary of \texttt{pred}; (f) expanded boundary of \texttt{gt}; (g) expanded boundary of \texttt{pred}; (h) pixel-wise multiplication of masks (d) and (g); (i) pixel-wise multiplication of masks (e) and (f)}
\label{fig:demo_loss}
\end{figure}

First of all, we present a working principle of the new loss in Fig.~\ref{fig:demo_loss}. In the upper row (see (b), (d) and (f)) we depict ground truth, where (b) is a binary mask for the ground truth segment, (d) is a binary mask of the \texttt{gt} boundary after applying (\ref{eq:gt_b}). Then we compute expanded boundary (f) using pooling operations (\ref{eq:gt_ext}). The same pipeline is applied to the predicted mask (color intensity represents the probability that pixel belongs to a foreground). After that we obtain \texttt{Precision map}, which represents pixel-wise multiplication of maps corresponding to (d) and (g), and \texttt{Recall map}, which is a multiplication of maps (e) and (f). \texttt{Precision} and \texttt{Recall} are normalized pixel-wise sums of these maps.

\begin{figure}[ht!]
\centering
\begin{tabular}{cccc}
  \includegraphics[width=27mm, height=66mm]{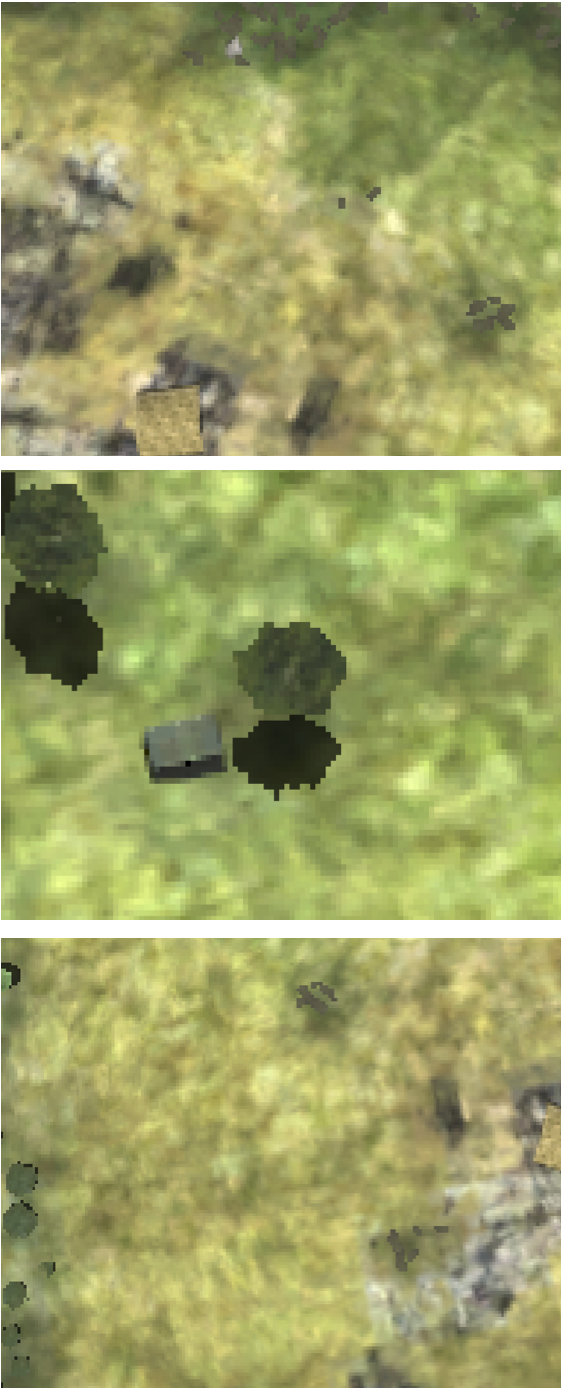} &
  \includegraphics[width=27mm]{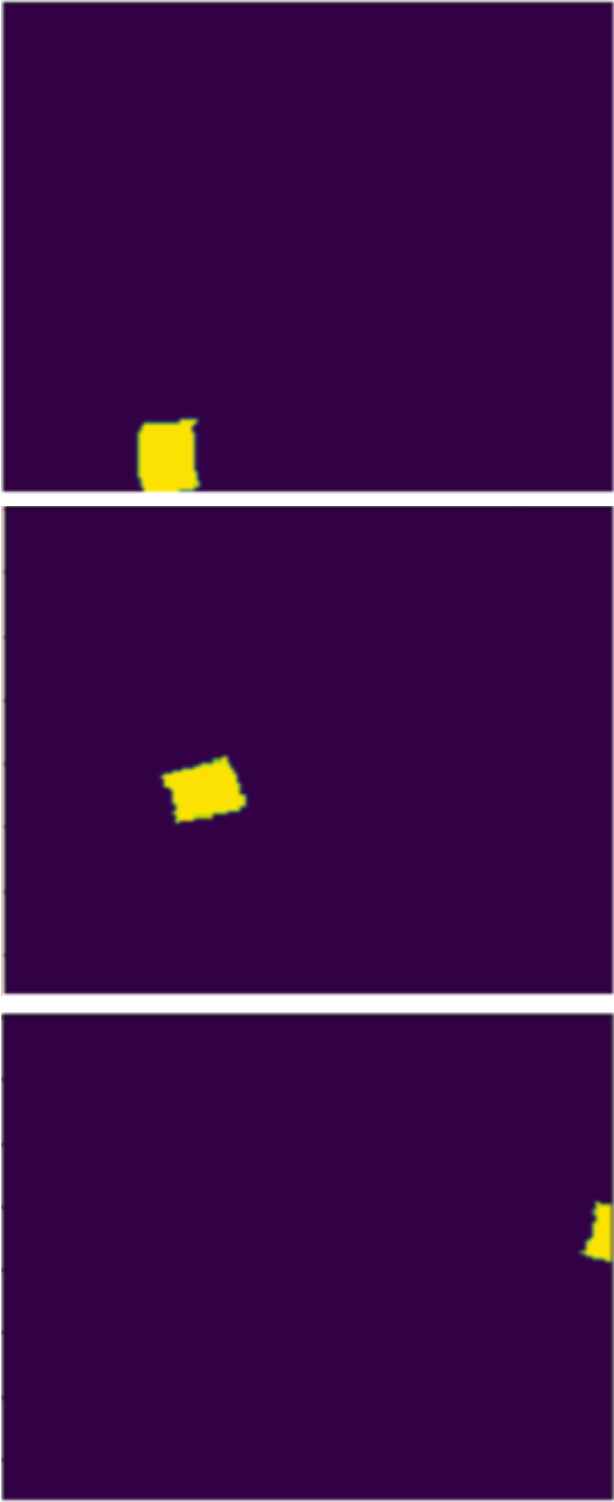} & \includegraphics[width=27mm]{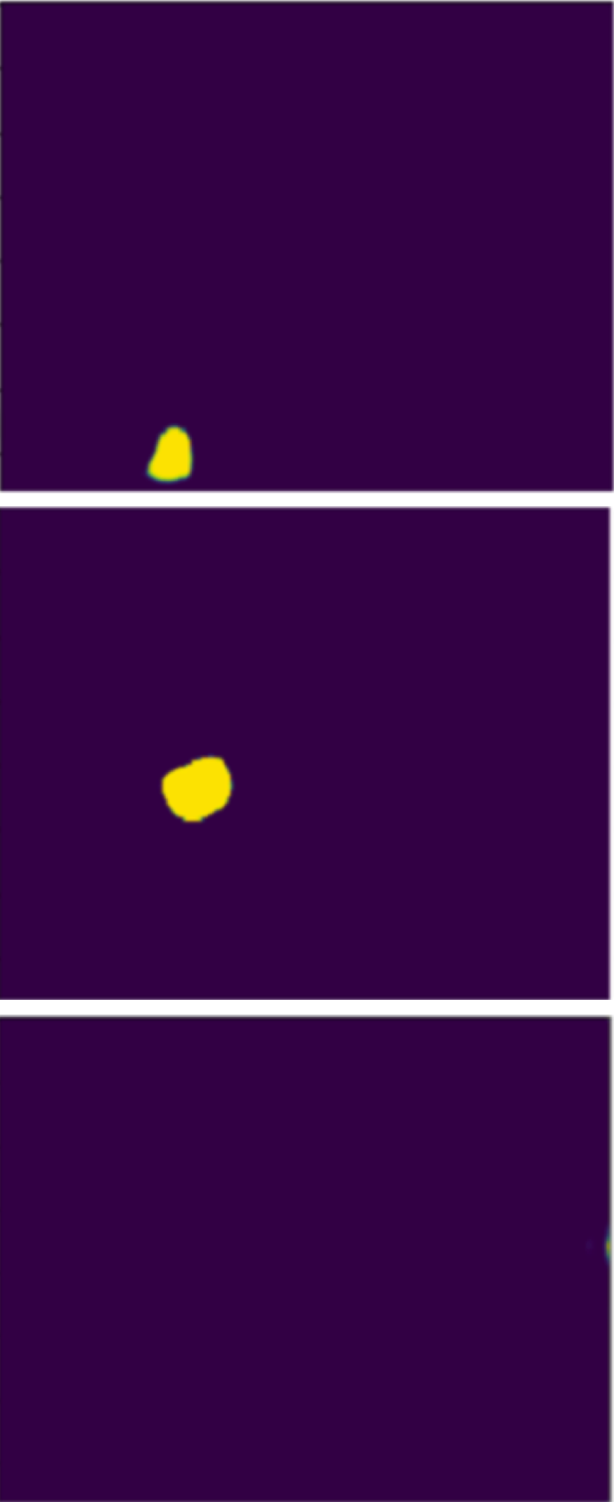} &  \includegraphics[width=27mm]{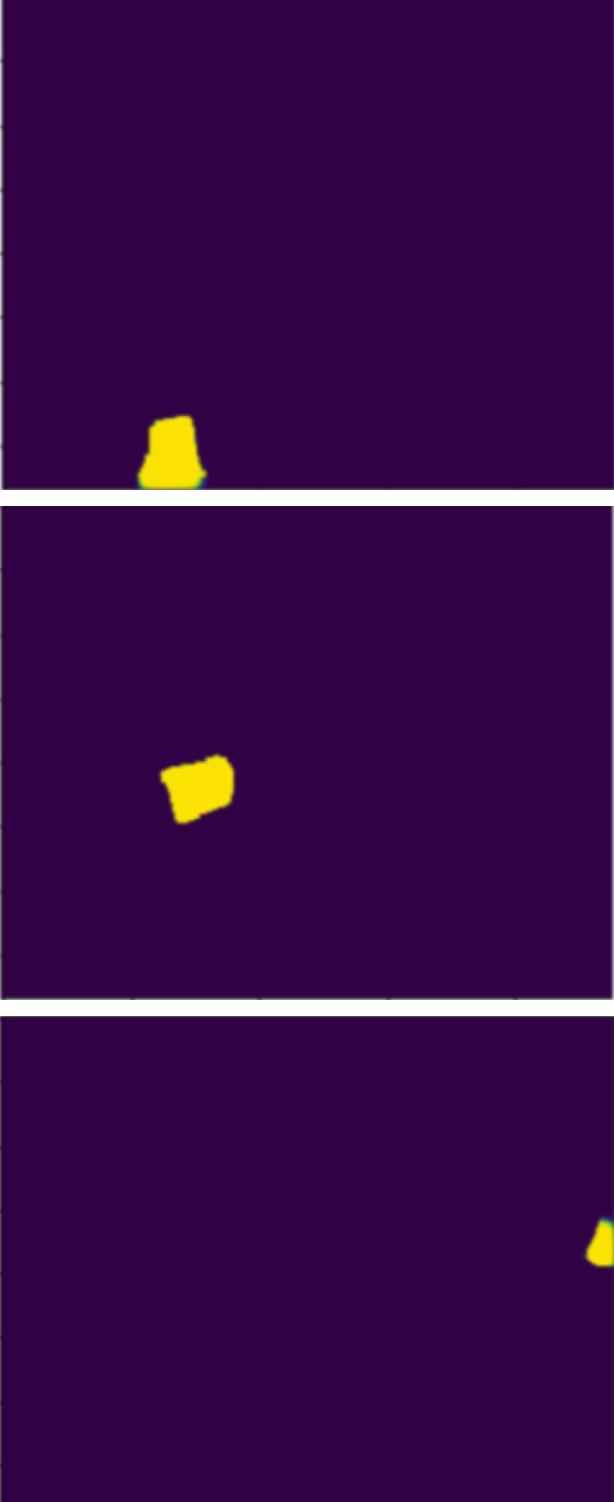}\\
(a)  & (b)  & (c) & (d) \\[6pt]
 \end{tabular}
\caption{(a) original orthophotos; (b) binary maps of ground truth segments; (c) predicted binary maps of segments ($L_{IoU}$ loss); (d) predicted binary maps of segments ($L_{BF_1, IoU}$ loss)}
\label{fig:aicd_examles}
\end{figure}

Some examples of how $L_{BF_1, IoU}$ loss outperforms $L_{IoU}$ loss are presented in Fig.~\ref{fig:aicd_examles}. It is clear that in comparison with $L_{IoU}$ training with $L_{BF_1, IoU}$ helps better delineate segment borders: they are very similar to ground truth boundaries and even meet at the corners in places where they should be. The second advantage is that the loss better handles edge effects. The third column clearly explains this kind of a problem when a building is located near the edge of the image. $L_{IoU}$ loss is unable to segment the building near the edge, however $L_{BF_1, IoU}$ does it accurately. Finally, the boundary loss converges faster on this synthetic dataset, and such behavior is expected not only for this simple data distribution. The intuition behind is that network keeps attention mainly on a boundary, which consists of fewer pixels in comparison with the entire segment.

\subsection{INRIA AIL dataset segmentation}\label{sec:Segmentation of INRIA AIL}

The dataset \cite{inria}  consists of 180 tiles (RGB) of 405 $km^2$ area in total with GSD of $0.3$ m. There are only two classes: \textit{building} and \textit{not building}. 

\begin{figure}[ht!]
\centering
\begin{minipage}[b]{.45\textwidth}
\includegraphics[width=60mm]{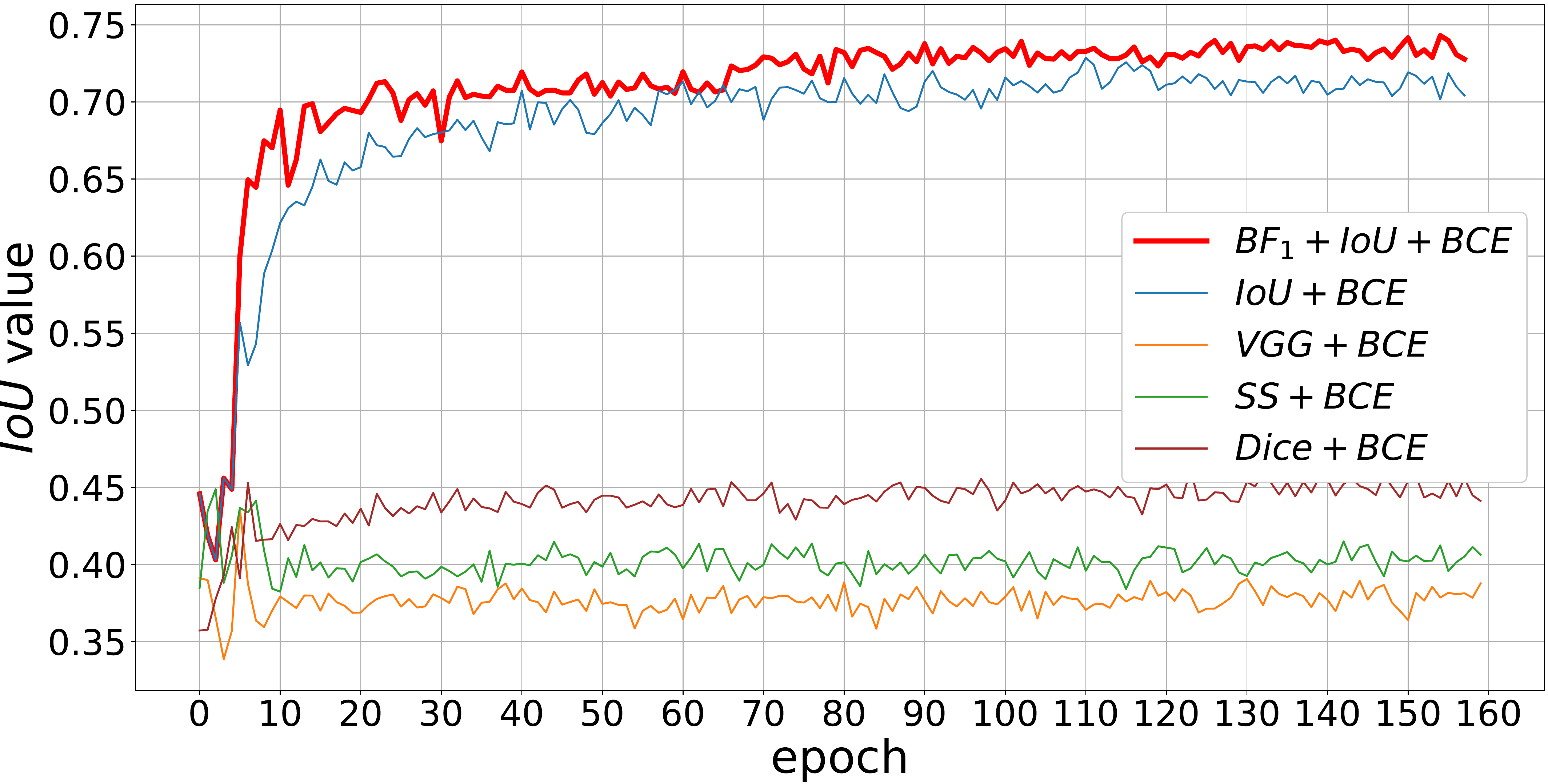}
\caption{Evolution of $IoU$ value during training for various loss functions (\mbox{INRIA AIL})}
\label{fig:inria_evolution}
\end{minipage}\qquad
\begin{minipage}[b]{.45\textwidth}
\includegraphics[width=60mm]{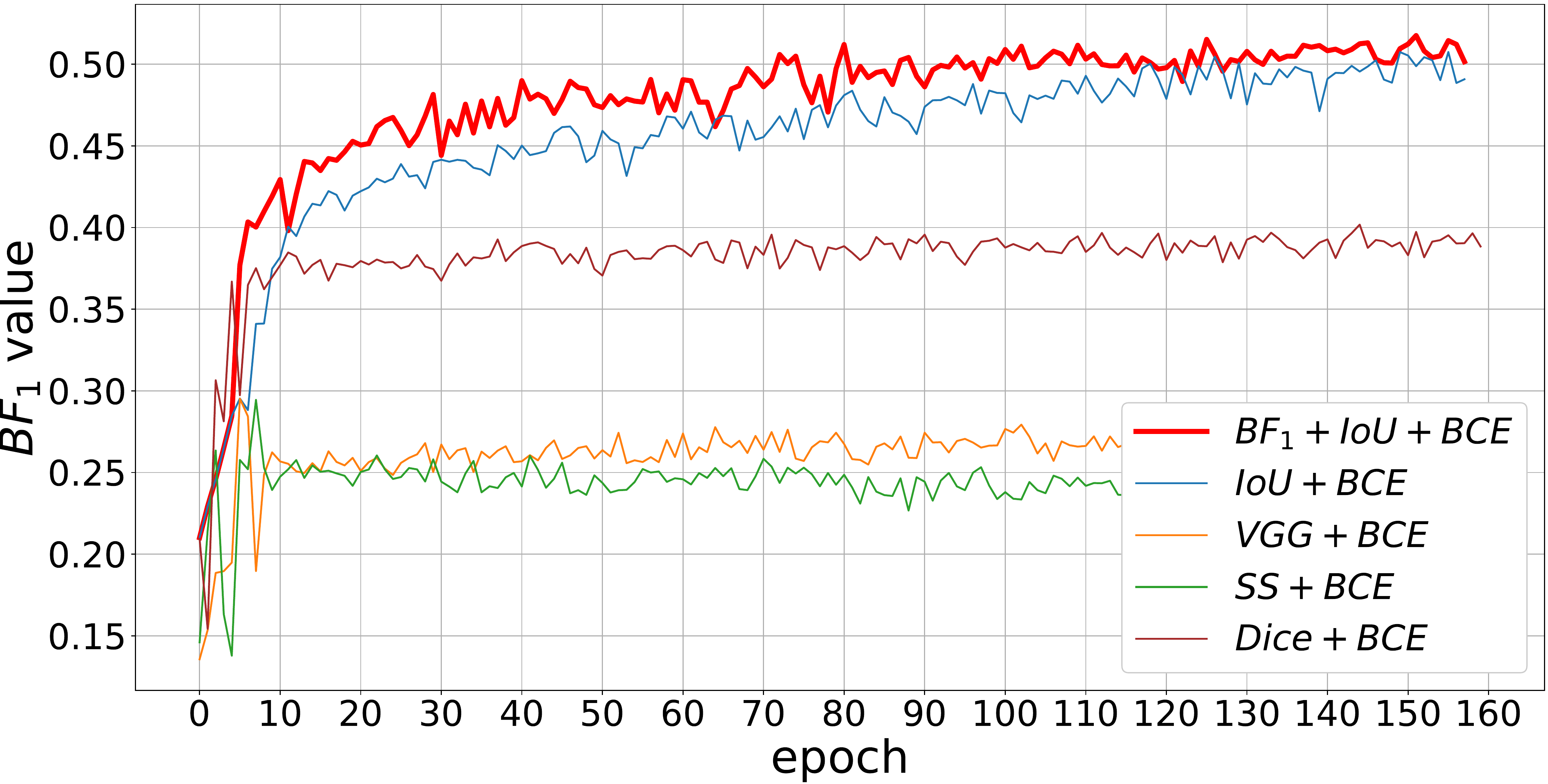}
\caption{Evolution of $BF_1$ value during training for various loss functions (\mbox{INRIA AIL})}
\label{fig:inria_bf}
\end{minipage}
\end{figure}

Training on this dataset is more complex than on AICD. As for preprocessing, very strong augmentations were applied including vertical, horizontal flips, hue/saturation adjustments, many kinds of noises and blurs. Every orthophoto was split into patches of size $512\times512$ and UNet with Inception-ResNet-v2 \cite{Szegedy2016Inceptionv4IA} as backbone was trained on these patches. After that, all predicted masks for every patch were merged into one mask of the whole orthophoto. Also, test-time augmentation (flips, reflections) was applied during the inference step. As for the training process, first, the network was trained from ImageNet \cite{imagenet} weights with a frozen encoder for three epochs, and the learning rate was set to $3\mathrm{e}{-3}$. After that all batch-normalization layers were unfrozen, and the learning rate was decreased to $1\mathrm{e}{-3}$. Then all layers were unfrozen and trained for 100 epochs with the learning rate of $1\mathrm{e}{-4}$. Finally, the network was trained for 60 epochs with the learning rate of $1\mathrm{e}{-5}$. We used Adam optimizer with default Keras \cite{chollet2015keras} settings. There were several experiments with different loss functions. As for $L_{IoU}$, $L_{Dice}$, $L_{SS}$ and $L_{VGG}$ losses, they did not require any adjustments while training: these losses were trained as a weighted sum with $BCE$ one by one. As for $L_{BF_1, IoU}$ loss, it requires an additional procedure for mini grid-search: after the 8th epoch for every 30 epochs and for every weight $w \in \{0.1, 0.3, 0.5, 0.7, 0.9\}$ in equation $(BCE + wL_{BF_1} + (1-w)L_{IoU})$ a network was trained. Then the best weight is chosen, and the process repeats. In Fig.~\ref{fig:inria_evolution} we see the evolution of $IoU$ metric while training with combinations of $BCE$ and $L_{IoU}$, $L_{Dice}$, $L_{SS}$, $L_{VGG}$, or $L_{BF_1, IoU}$. $BF_1$ is also a very important metric for remote sensing image segmentation (see Fig.~\ref{fig:inria_bf}).

Finally, from figures, we see that the final segmentation is better than the baseline model trained with $L_{IoU}$ in terms of $IoU$ and $BF_1$. Now we want to understand what is better when training with the boundary loss. Below we provide an example comparing two segmentations.

\begin{figure}[ht!]
\centering
\begin{minipage}[b]{.45\textwidth}
\begin{tabular}{cc}
\includegraphics[width=29.5mm, height=28mm]{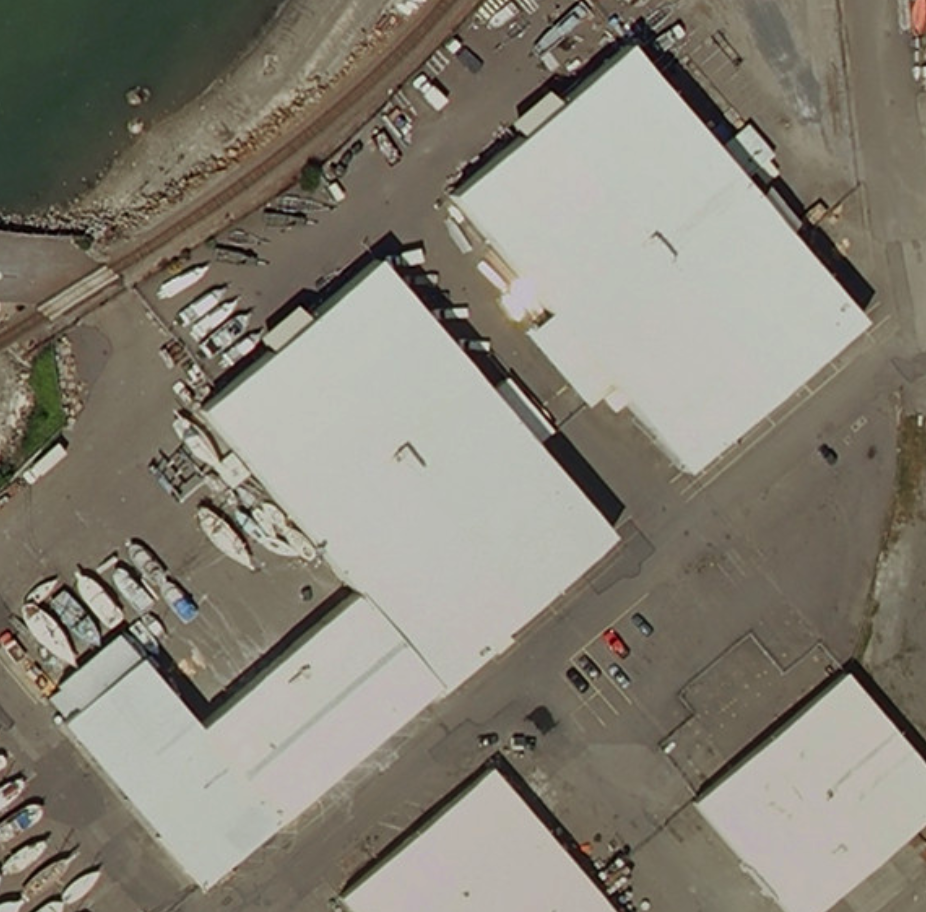} & \\
(a) & \\[6pt]
\includegraphics[width=29.5mm]{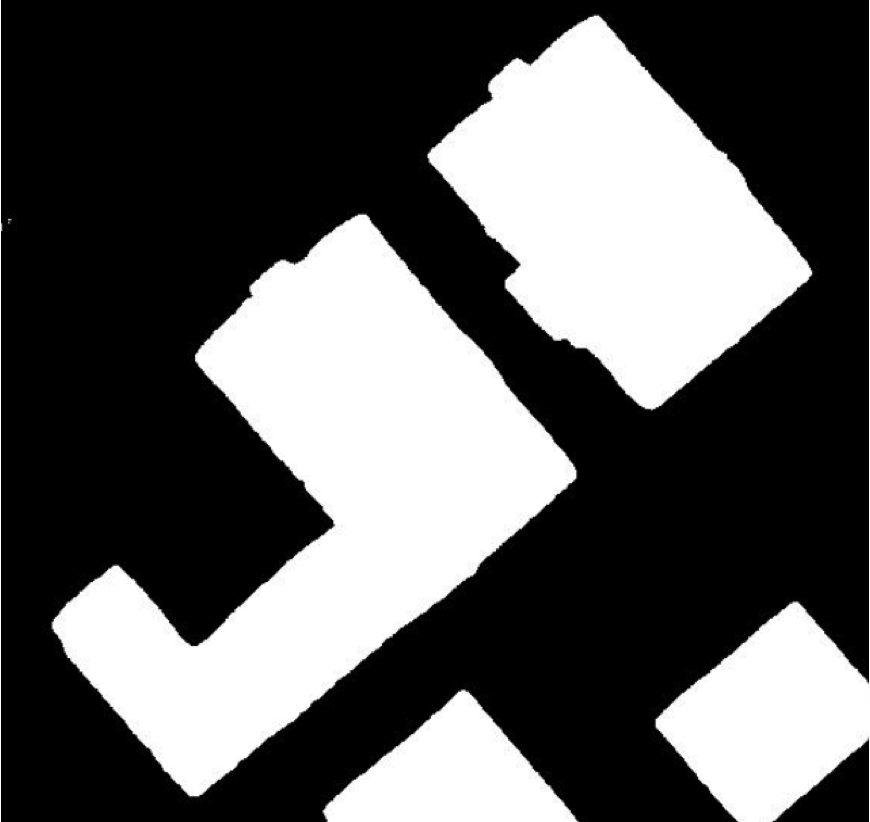} &   \includegraphics[width=29.5mm]{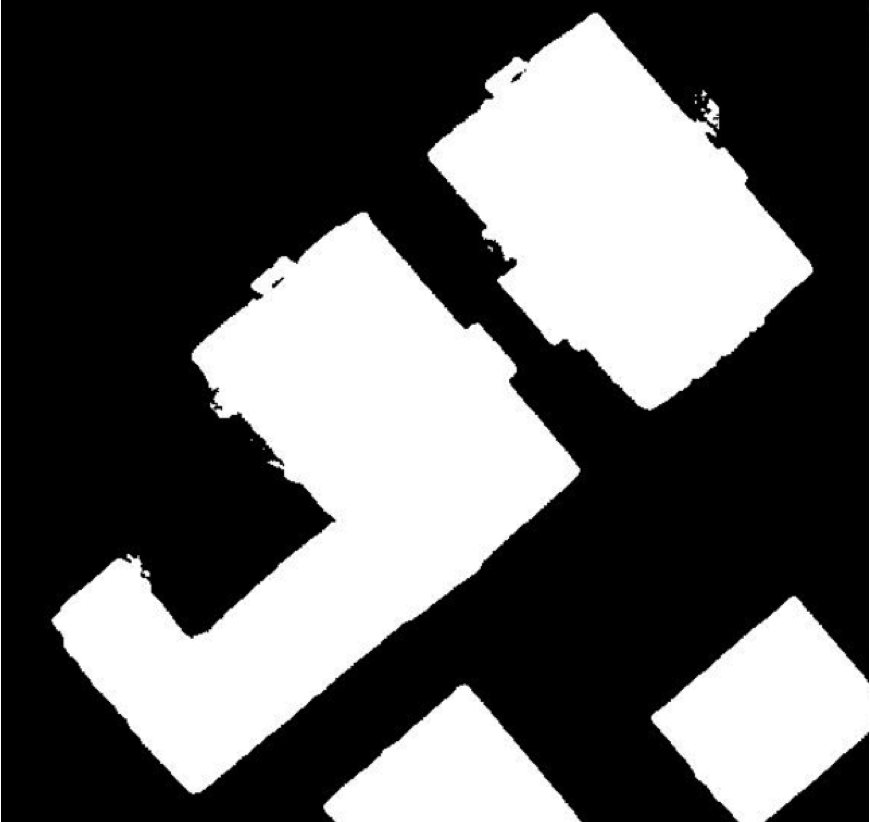} \\
(b)  & (c)  \\[6pt]
\end{tabular}
\vspace{2.5mm}
\caption{(a) original orthophoto; (b) predicted segmentation trained with $L_{IoU}$; (c) predicted segmentation trained with $L_{BF_1, IoU}$}
\label{fig:inria_ex1}\end{minipage}\qquad
\begin{minipage}[b]{.45\textwidth}
\begin{tabular}{cc}
 \includegraphics[width=28mm]{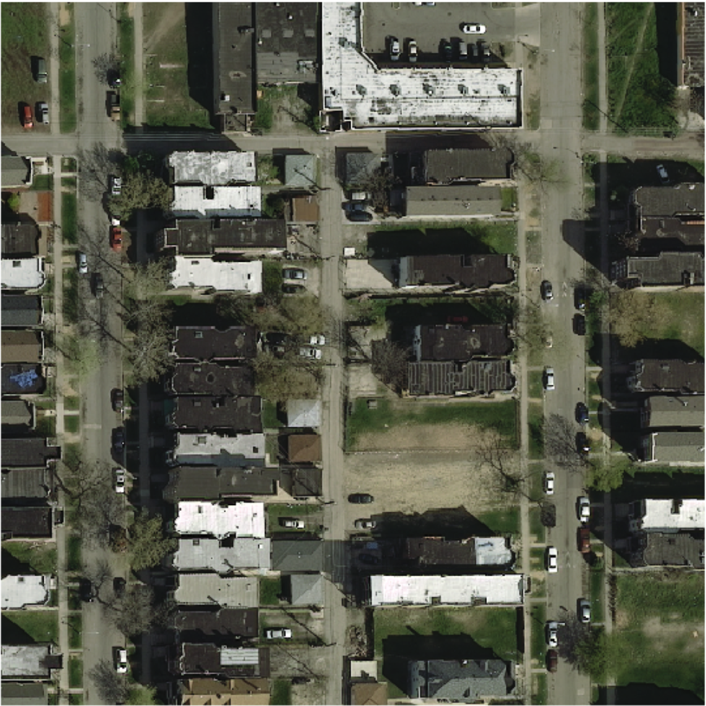} &   \includegraphics[width=28mm]{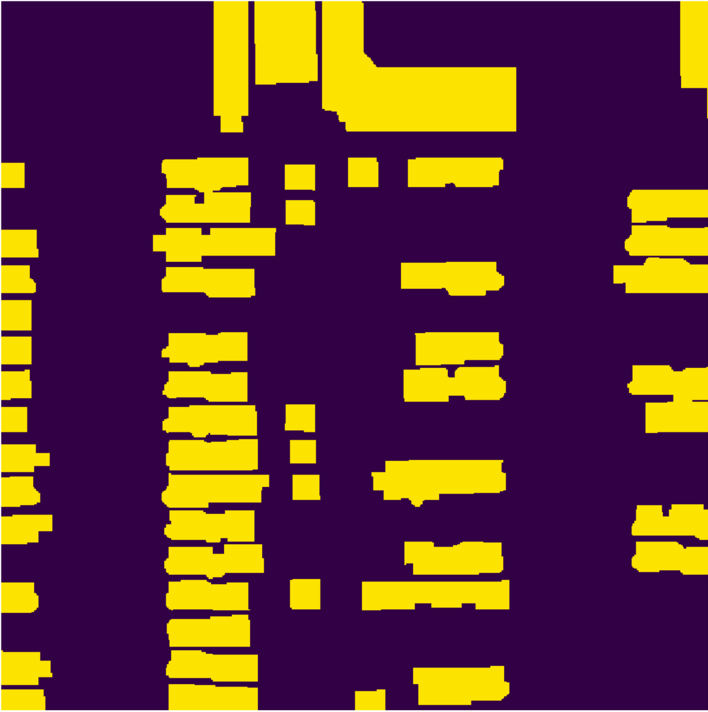} \\
(a)  & (b)  \\[6pt]
\includegraphics[width=28mm]{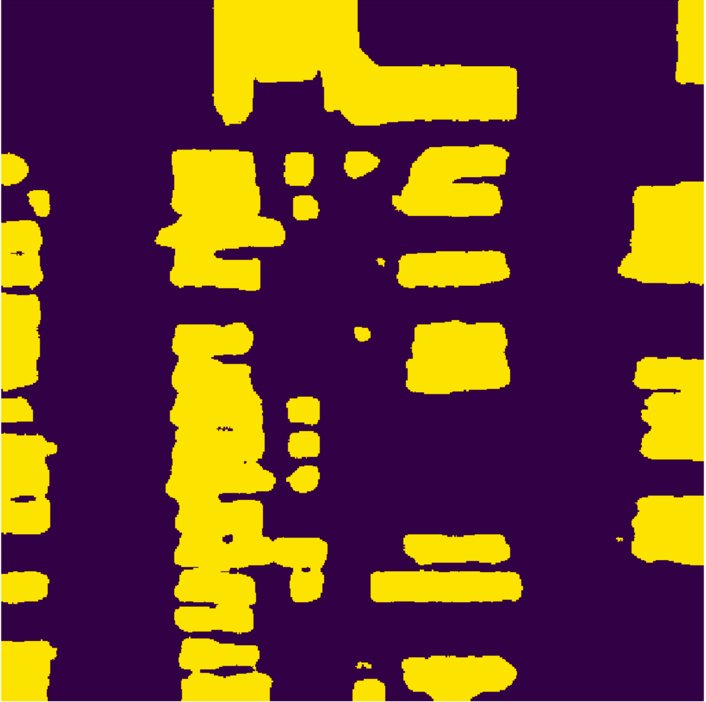} &   \includegraphics[width=28mm]{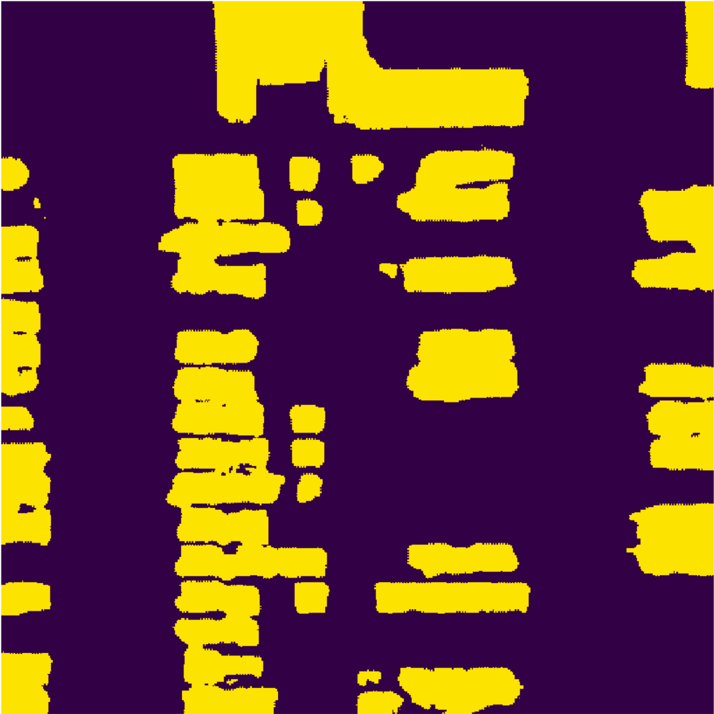} \\
(c)  & (d)  \\[6pt]
\end{tabular}
\caption{(a) original orthophoto; (b) ground truth segmentation \texttt{(gt)}; (c) predicted segmentation trained with $L_{IoU}$; (d) predicted segmentation trained with $L_{BF_1, IoU}$}
\label{fig:inria_ex2}
\end{minipage}
\end{figure}

In Fig.~\ref{fig:inria_ex1} on the right we see more accurate shapes of edges and corners.  The important feature, which follows from Fig. \ref{fig:inria_ex2}, is that several buildings are standing in one row with a distance between adjacent constructions about 5 pixels. In comparison with $L_{IoU}$, $L_{BF_1, IoU}$ our loss managed to separate instances of buildings much better. 

\subsection{ISPRS Potsdam dataset segmentation}\label{sec:Segmentation of ISPRS Potsdam}

The dataset \cite{isprs} contains 38 patches, each consisting of a true orthophoto extracted from the larger mosaic. There are five channels (RGB+NIR+DSM) with GSD  $0.2$ m for each channel. As a model, UNet was trained with very strong augmentations similar to that used for INRIA AIL. We used test-time augmentation (flips, reflections), set the learning rate to $1\mathrm{e}{-3}$ and divided it by a factor of $10$ if for $10$ epochs there was no improvement in validation loss. We trained a neural network on patches of orthophotos with Adam optimizer. 

\begin{figure}[ht!]
\centering
\begin{minipage}[b]{.45\textwidth}
\includegraphics[width=60mm]{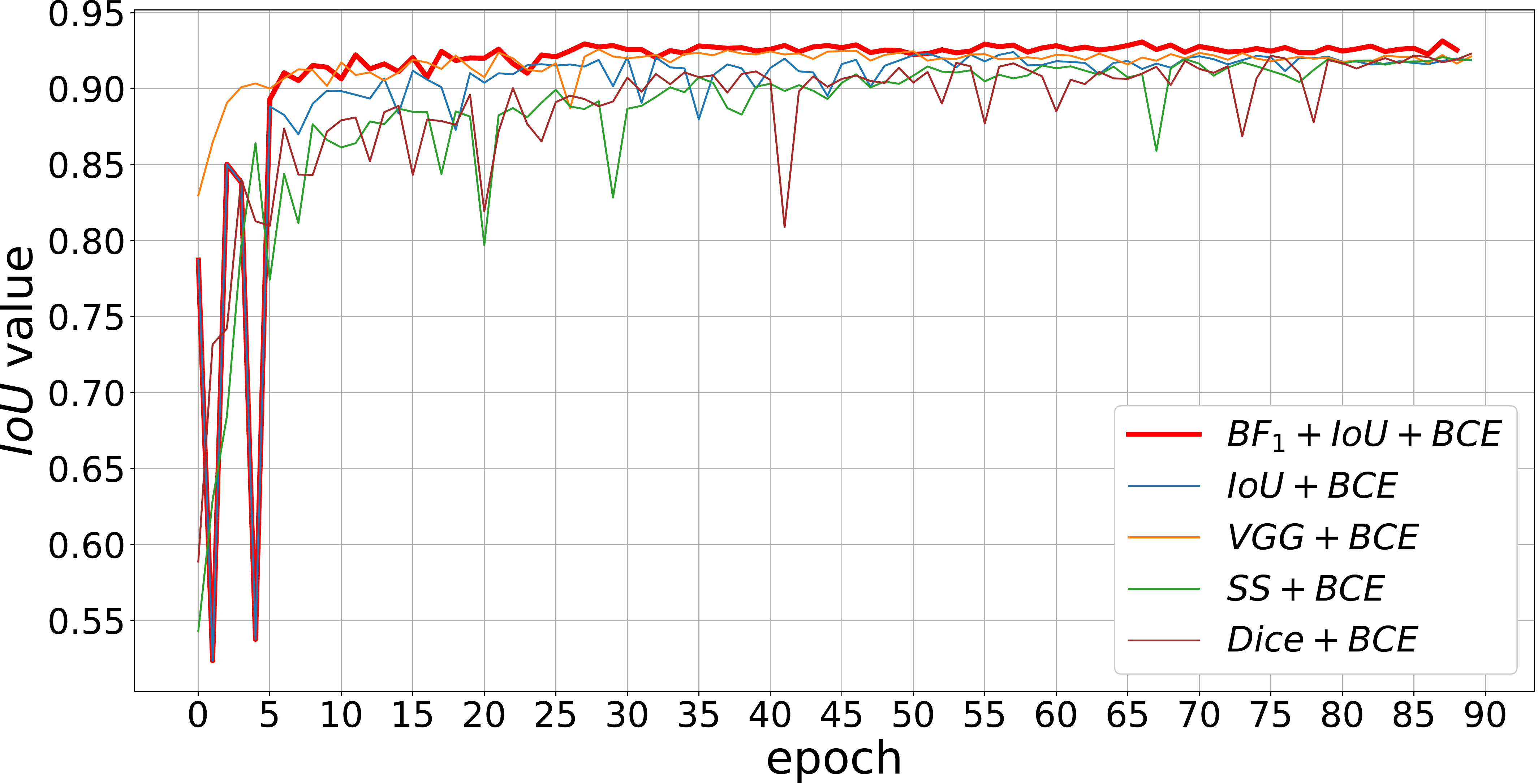}
\caption{Evolution of $IoU$ value during training for various loss functions (\mbox{ISPRS Potsdam})}
\label{fig:isprs_evolution}
\end{minipage}\qquad
\begin{minipage}[b]{.45\textwidth}
\includegraphics[width=60mm]{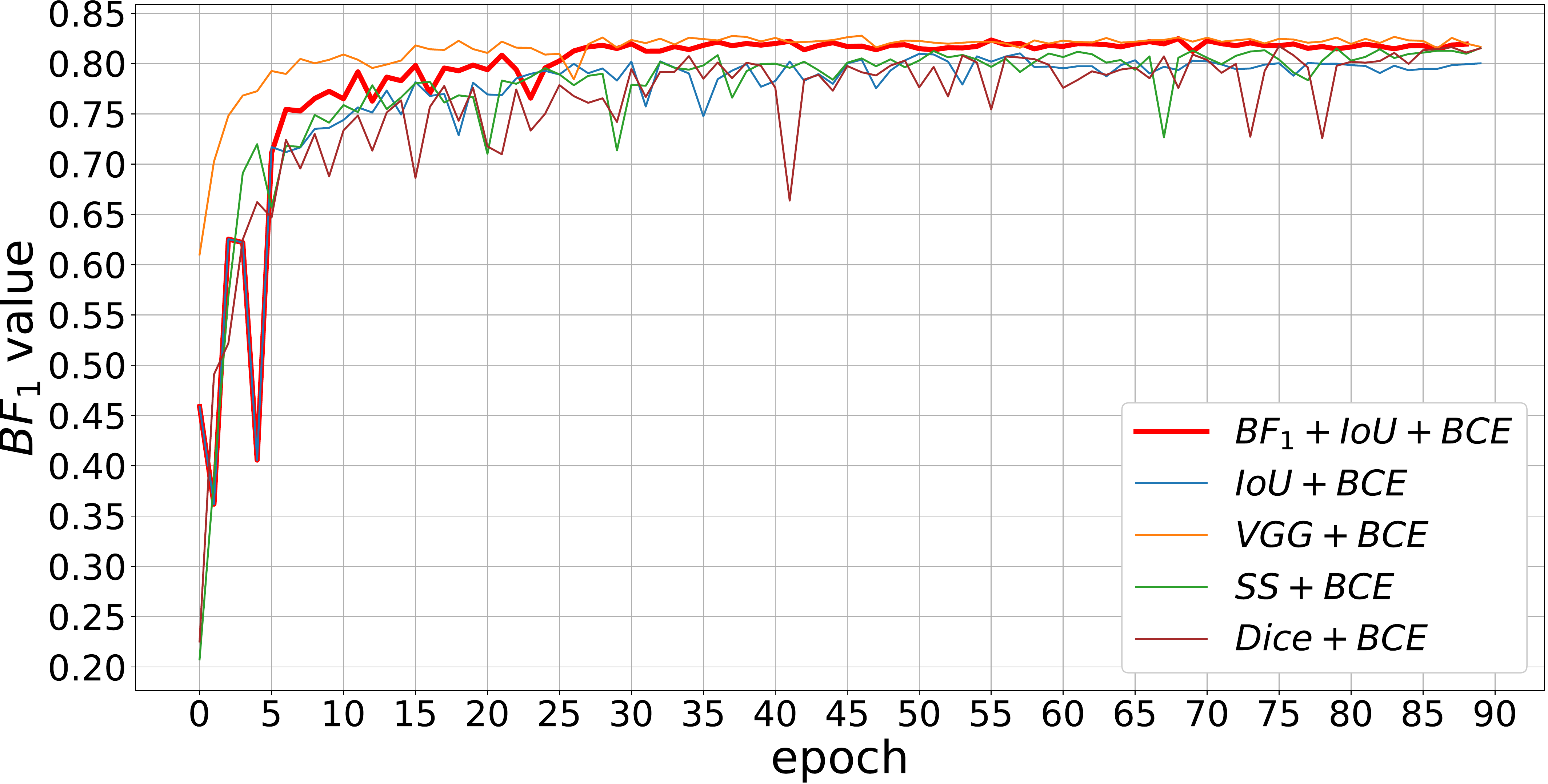}
\caption{Evolution of $BF_1$ value during training for various loss functions (\mbox{ISPRS Potsdam})}
\label{fig:isprs_bf}
\end{minipage}
\end{figure}

In Fig.~\ref{fig:isprs_evolution} and Fig.~\ref{fig:isprs_bf} we see the evolution of $IoU$ and $BF_1$ metric while training with combinations of $BCE$ and $L_{IoU}$, $L_{Dice}$, $L_{SS}$, $L_{VGG}$, $L_{BF_1, IoU}$. Here we see slightly faster convergence of $IoU$ for $L_{BF_1, IoU}$, the final segmentation is 1-2\% better than baseline $(BCE+L_{IoU})$. Also there is an advantage that training with $L_{BF_1, IoU}$ loss is much less noisy.

\begin{figure}[ht!]
\centering
\includegraphics[width=75mm]{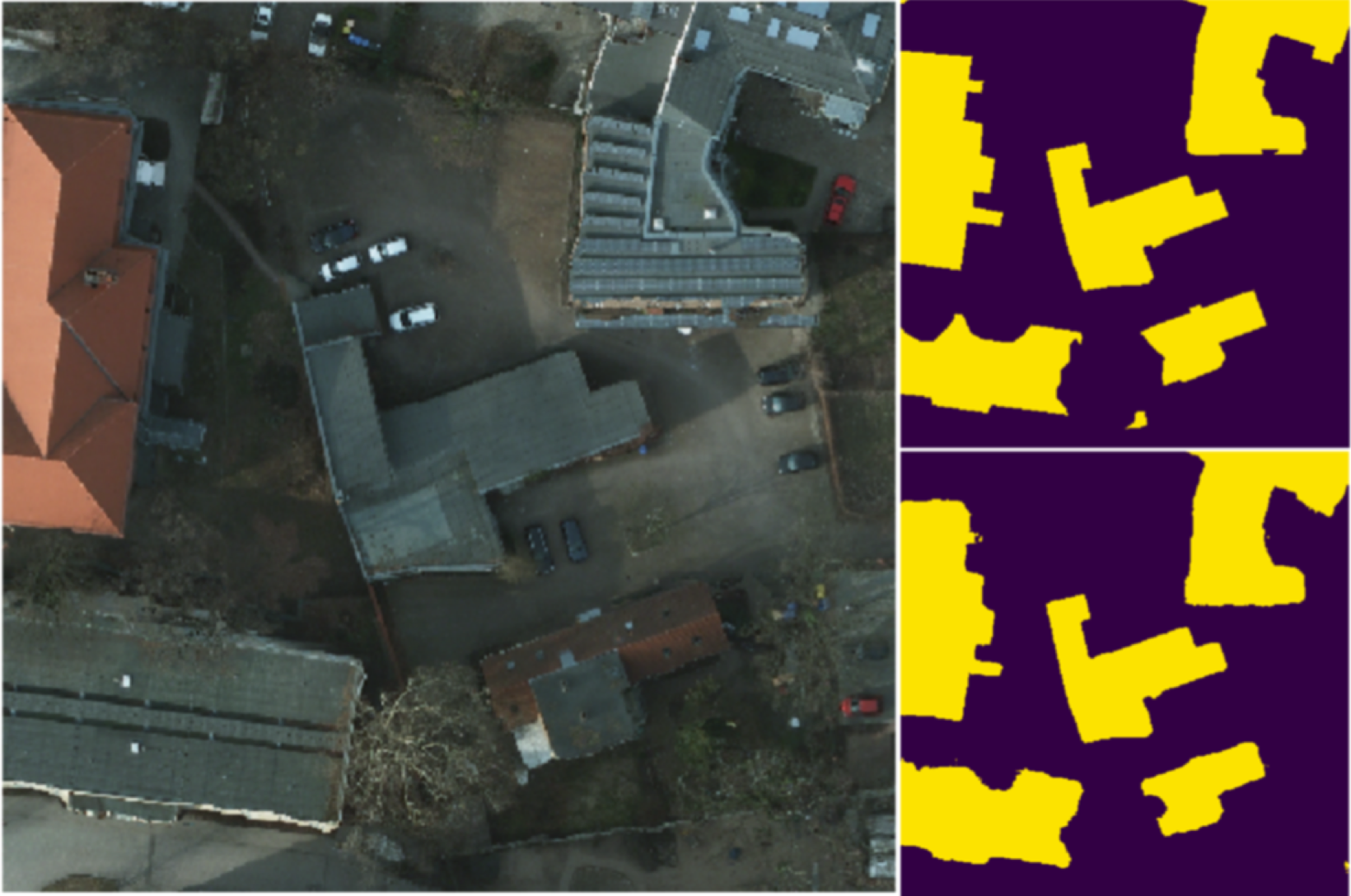}
\caption{Example of segmentation (loss --- $L_{IoU}$). \textbf{Left}: original orthophoto, \textbf{top right}: ground truth segmentation, \textbf{bottom right}: predicted segmentation}
\label{fig:isprs_ex1}
\end{figure}

\begin{figure}[ht!]
\centering
\includegraphics[width=75mm]{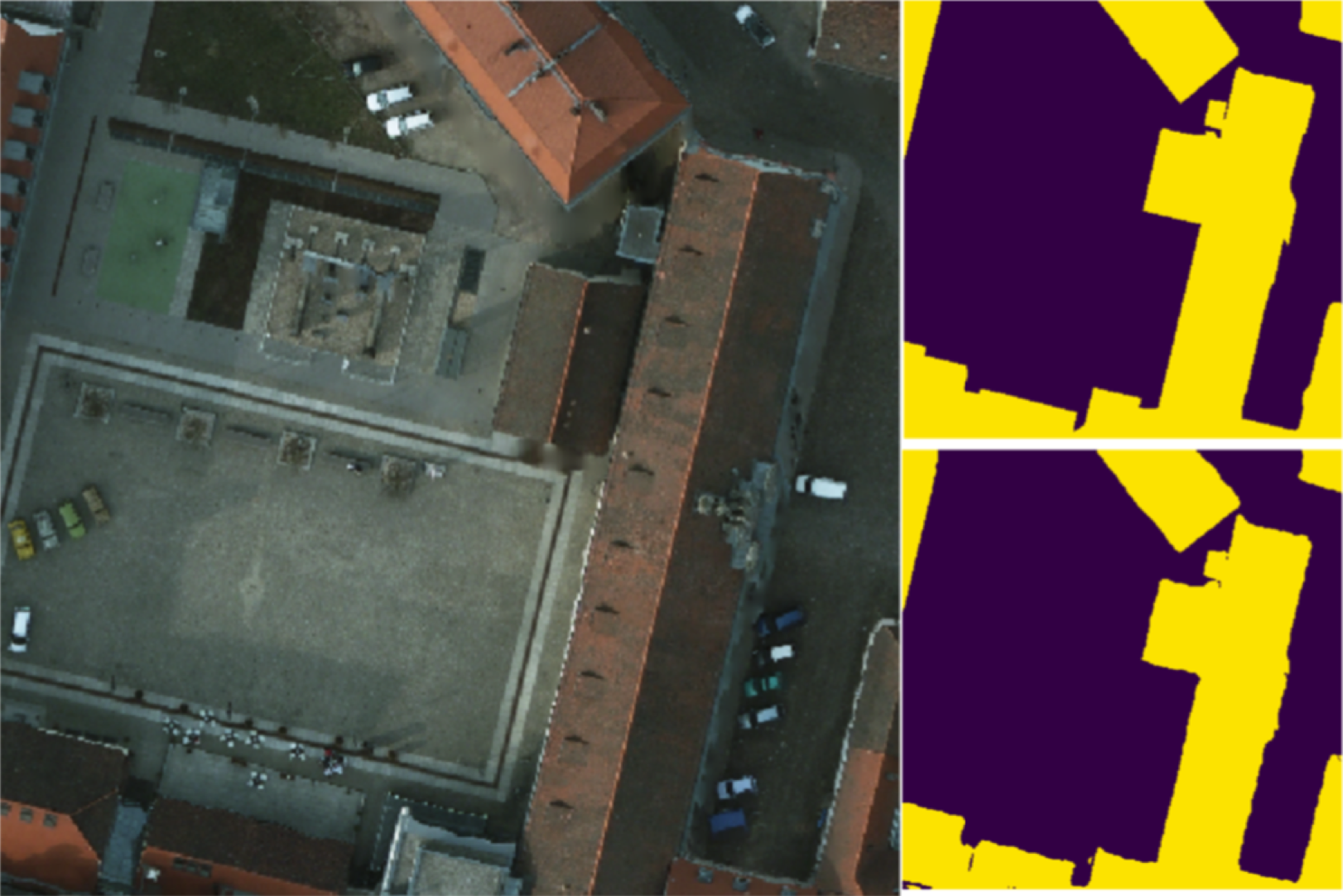}
\caption{Example of segmentation (loss --- $L_{BF_1, IoU}$). \textbf{Left}: original orthophoto, \textbf{top right}: ground truth segmentation, \textbf{bottom right}: predicted segmentation}
\label{fig:isprs_ex2}
\end{figure}

Even though $IoU$ value is the best with $L_{BF_1, IoU}$ training, boundary metric converges faster with VGG loss; authors of the original paper mentioned that this loss function is also efficient for boundaries delineation. In Fig.~\ref{fig:isprs_ex1} and Fig.~\ref{fig:isprs_ex2} there are several examples segmented with $L_{IoU}$ and the boundary loss.

Direct $IoU$ loss (Fig.~\ref{fig:isprs_ex1}) has poorer performance on edges of segments and complicated shapes, whereas boundary loss (Fig.~\ref{fig:isprs_ex2}) on the right retrieved almost all tricky shapes. 

In Table \ref{table:jaccard1} and Table \ref{table:jaccard2} we provide final results of different losses in the task of binary segmentation of buildings for different models and datasets. Backbones for comparison in these tables are chosen is such way that they showed the best performance over all we have tried (VGG16, VGG19, ResNet34, DenseNet121, DenseNet169, Inception-ResNet-v2).

\begin{table}[ht]
\centering 
\begin{tabular}{c c c c} 
\hline\hline 
& AICD & ISPRS Potsdam & INRIA AIL \\ [0.5ex] 
& ConvNN & ResNet34 & Inc.-ResNet-v2 \\ [0.5ex] 
\hline
$L_{BF_1, IoU}$ & \textbf{86.91} & \textbf{93.85} & \textbf{74.30} \\ 
$L_{IoU}$ & 85.66 & 92.56 & 71.91 \\
$L_{SS, IoU}$ & $61.14^*$ & 92.47 & 43.66 \\
$L_{VGG, IoU}$ & $85.76^*$ & 92.53 & 45.89 \\
$L_{Dice, IoU}$ & $85.13^*$ & 92.61 & 44.90 \\ [1ex] 
\hline \\
\end{tabular}
\caption{$IoU$ values for different datasets and loss functions. ConvNN is a not very deep neural network with 5 convolutional blocks in both the encoder and the decoder. For ISPRS dataset UNet model with ResNet34 backbone was used, for INRIA dataset -- UNet with Inception-ResNet-v2 backbone}
\label{table:jaccard1} 
\end{table}
\begin{table}[ht]
\centering 
\begin{tabular}{c c c c} 
\hline\hline 
& AICD & ISPRS Potsdam & INRIA AIL \\ [0.5ex] 
& ConvNN & ResNet34 & Inc.-ResNet-v2 \\ [0.5ex] 
\hline
$L_{BF_1, IoU}$ & \textbf{71.51} & 82.48 & \textbf{51.75} \\ 
$L_{IoU}$ & 68.01 & 81.02 & 50.75 \\
$L_{SS, IoU}$ & $20.74^*$ & 81.55 & 29.53 \\
$L_{VGG, IoU}$ & $69.41^*$ & \textbf{82.77} & 40.18 \\
$L_{Dice, IoU}$ & $70.49^*$ & 81.86 & 29.45 \\ [1ex]
\hline \\
\end{tabular}
\caption{$BF_1$ values for different datasets and loss functions. For ISPRS dataset UNet model with \textit{ResNet34} backbone was used, for INRIA dataset -- UNet with \textit{Inception-ResNet-v2} backbone}
\label{table:jaccard2} 
\end{table}

Finally, here are some constraints and recommendations:
\begin{enumerate}
\item
It is better to use boundary loss only for segments with precise edges and corners such as buildings, roads etc.
\item Let us discuss the choice of parameters $\theta_0$ and  $\theta$ (\ref{eq:gt_b}) and (\ref{eq:gt_ext}) respectively. The best choice for the parameter $\theta_0$ is as less as possible, but at the same time it must be possible to extract solid boundaries of segments.  The next constraint is that $\theta_0$ and $\theta$ should be smaller than minimum distance between segments, otherwise the performance of segmenting current element is affected by other segments because of overlapped expanded boundaries. Via the trial and error process we set $\theta_0$  to $3$ and $\theta$ to $5$-$7$ as a proper choice, because theses values deliver the most accurate boundaries in all experiments.
\end{enumerate}

\section{Conclusions}\label{CONCLUSIONS}

In this work, we introduced a novel loss function to encourage a neural network better taking into account segment boundaries. As motivation, we applied this approach to a synthetic dataset of binary segmentation and after that proved its validity on real-world data. Our experimental results demonstrate that optimizing $IoU$ assisted by $BF_1$ metric surrogate leads to  better performance and more accurate boundaries delineation compared to using other loss functions. As for future work, we would like to extend our approach to handle multi-class semantic segmentation. We are going to elaborate on this approach by increasing its computational capabilities with large scale sparse convolutional neural networks \cite{3DCNN2018}, using a new loss function, specially tailored for imbalanced classification \cite{Imbalance2019,Imbalanced2015}, utilizing an approach for combining multi-modal data through CNNs features aggregation \cite{Multispectral2018} and imposing a confidence measure on top of the segmentation model based on the non-parametric conformal approach \cite{kNN2017,ConformalAD2015,ConformalMartingales2017}.

\bibliographystyle{splncs04}
\bibliography{references}

\begin{thebibliography}{10}
\providecommand{\url}[1]{\texttt{#1}}
\providecommand{\urlprefix}{URL }
\providecommand{\doi}[1]{https://doi.org/#1}

\bibitem{crf}
{Alam}, F.I., {Zhou}, J., {Liew}, A.W., {Jia}, X., {Chanussot}, J., {Gao}, Y.:
  Conditional random field and deep feature learning for hyperspectral image
  classification. IEEE Transactions on Geoscience and Remote Sensing
  \textbf{57}(3),  1612--1628 (2019). \doi{10.1109/TGRS.2018.2867679}

\bibitem{Badrinarayanan2016SegNetAD}
Badrinarayanan, V., Kendall, A., Cipolla, R.: Segnet: A deep convolutional
  encoder-decoder architecture for image segmentation. IEEE Transactions on
  Pattern Analysis and Machine Intelligence  \textbf{39},  2481--2495 (2016)

\bibitem{lovasz}
Berman, M., Triki, A.R., Blaschko, M.B.: The lovasz-softmax loss: A tractable
  surrogate for the optimization of the intersection-over-union measure in
  neural networks. 2018 IEEE/CVF Conference on Computer Vision and Pattern
  Recognition pp. 4413--4421 (2018)

\bibitem{Multispectral2018}
Burnaev, E., Cichocki, A., Osin, V.: Fast multispectral deep fusion networks.
  Bull. Pol. Ac.: Tech.  \textbf{66}(4),  875--880 (2018)

\bibitem{Imbalanced2015}
Burnaev, E., Erofeev, P., Papanov, A.: Influence of resampling on accuracy of
  imbalanced classification. In: Eighth International Conference on Machine
  Vision, 987525 (December 8, 2015). Proc SPIE, vol.~9875 (2015).
  \doi{10.1117/12.2228523}

\bibitem{DeepLab}
{Chen}, L., {Papandreou}, G., {Kokkinos}, I., {Murphy}, K., {Yuille}, A.L.:
  Deeplab: Semantic image segmentation with deep convolutional nets, atrous
  convolution, and fully connected crfs. IEEE Transactions on Pattern Analysis
  and Machine Intelligence  \textbf{40}(4),  834--848 (2018).
  \doi{10.1109/TPAMI.2017.2699184}

\bibitem{chollet2015keras}
Chollet, F., et~al.: Keras. \url{https://github.com/fchollet/keras} (2015)

\bibitem{cityscapes}
Cordts, M., Omran, M., Ramos, S., Rehfeld, T., Enzweiler, M., Benenson, R.,
  Franke, U., Roth, S., Schiele, B.: The cityscapes dataset for semantic urban
  scene understanding. In: Proc. of the IEEE Conference on Computer Vision and
  Pattern Recognition (CVPR) (2016)

\bibitem{bf1}
Csurka, G., Larlus, D., Perronnin, F.: What is a good evaluation measure for
  semantic segmentation? IEEE PAMI  \textbf{26},  1--11 (2004)

\bibitem{imagenet}
{Deng}, J., {Dong}, W., {Socher}, R., {Li}, L., and: Imagenet: A large-scale
  hierarchical image database. In: 2009 IEEE Conference on Computer Vision and
  Pattern Recognition. pp. 248--255 (2009)

\bibitem{pascal-voc-2012}
Everingham, M., Van~Gool, L., Williams, C.K.I., Winn, J., Zisserman, A.: The
  {PASCAL} {V}isual {O}bject {C}lasses {C}hallenge 2012 {(VOC2012)} {R}esults.
  www.pascal-network.org/challenges/VOC/voc2012/workshop/index.html

\bibitem{bf2}
Fernandez-Moral, E., Martins, R., Wolf, D., Rives, P.: {A new metric for
  evaluating semantic segmentation: leveraging global and contour accuracy}.
  In: {Workshop on Planning, Perception and Navigation for Intelligent
  Vehicles, PPNIV17}. Vancouver, Canada (Sep 2017),
  \url{https://hal.inria.fr/hal-01581525}

\bibitem{cch19}
Ignatiev, V., Trekin, A., Lobachev, V., Potapov, G., Burnaev, E.: Targeted
  change detection in remote sensing images. In: Eleventh International
  Conference on Machine Vision (ICMV 2018); 110412H (2019). Proc. SPIE, vol.
  11041 (2019). \doi{doi.org/10.1117/12.2523141}

\bibitem{kNN2017}
Ishimtsev, V., Bernstein, A., Burnaev, E., Nazarov, I.: Conformal k-nn anomaly
  detector for univariate data streams. In: Proc. of 6th Workshop COPA. PRML,
  vol.~60, pp. 213--227. PMLR (2017)

\bibitem{alexnet}
Krizhevsky, A., Sutskever, I., E.~Hinton, G.: Imagenet classification with deep
  convolutional neural networks. Advances in neural information processing
  systems pp. 1097--1105 (2012)

\bibitem{inria}
{Maggiori}, E., {Tarabalka}, Y., {Charpiat}, G., {Alliez}, P.: Can semantic
  labeling methods generalize to any city? the inria aerial image labeling
  benchmark. In: 2017 IEEE International Geoscience and Remote Sensing
  Symposium (IGARSS). pp. 3226--3229 (2017). \doi{10.1109/IGARSS.2017.8127684}

\bibitem{vgg}
Mosinska, A., Marquez-Neila, P., Kozinski, M., Fua, P.: Beyond the pixel-wise
  loss for topology-aware delineation. pp. 3136--3145 (06 2018).
  \doi{10.1109/CVPR.2018.00331}

\bibitem{neuro}
Nagendar, G., Singh, D., Balasubramanian, V.N., Jawahar, C.V.: Neuro-iou:
  Learning a surrogate loss for semantic segmentation. In: British Machine
  Vision Conference 2018, {BMVC} 2018, Northumbria University, Newcastle, UK,
  September 3-6, 2018. pp. 278--289 (2018),
  \url{http://bmvc2018.org/contents/papers/1055.pdf}

\bibitem{3DCNN2018}
Notchenko, A., Kapushev, Y., Burnaev, E.: Large-scale shape retrieval with
  sparse 3d convolutional neural networks. In: Analysis of Images, Social
  Networks and Texts. pp. 245--254. Springer (2018)

\bibitem{RSdamage2018}
Novikov, G., Trekin, A., Potapov, G., Ignatiev, V., Burnaev, E.: Satellite
  imagery analysis for operational damage assessment in emergency situations.
  In: Abramowicz, W., Paschke, A. (eds.) Business Information Systems -- 21st
  International Conference, {BIS} 2018, Berlin, Germany, July 18-20, 2018,
  Proceedings. Lecture Notes in Business Information Processing, vol.~320, pp.
  347--358. Springer (2018)

\bibitem{nowozin}
Nowozin, S.: Optimal decisions from probabilistic models: The
  intersection-over-union case. 2014 IEEE Conference on Computer Vision and
  Pattern Recognition pp. 548--555 (2014)

\bibitem{Pohlen2017FullResolutionRN}
Pohlen, T., Hermans, A., Mathias, M., Leibe, B.: Full-resolution residual
  networks for semantic segmentation in street scenes. 2017 IEEE Conference on
  Computer Vision and Pattern Recognition (CVPR) pp. 3309--3318 (2017)

\bibitem{unet}
Ronneberger, O., Fischer, P., Brox, T.: U-net: Convolutional networks for
  biomedical image segmentation. In: Navab, N., Hornegger, J., Wells, W.M.,
  Frangi, A.F. (eds.) Medical Image Computing and Computer-Assisted
  Intervention -- MICCAI 2015. pp. 234--241. Springer International Publishing,
  Cham (2015)

\bibitem{isprs}
Rottensteiner, F., Sohn, G., Jung, J., Gerke, M., Bailard, C., Benitez, S.,
  Breitkopf, U.: The isprs benchmark on urban object classification and 3d
  building reconstruction. In: Shortis, M., Paparoditis, N., Mallett, C. (eds.)
  ISPRS 2012 Proceedings of the XXII ISPRS Congress : Imaging a Sustainable
  Future, 25 August - 01 September 2012, Melbourne, Australia. Peer reviewed
  Annals, Volume I-7, 2012. pp. 293--298. International Society for
  Photogrammetry and Remote Sensing (ISPRS) (8 2012)

\bibitem{ConformalAD2015}
Safin, A., Burnaev, E.: Conformal kernel expected similarity for anomaly
  detection in time-series data. Adv. in Systems Sci.and Appl.  \textbf{17}(3),
   22--33 (2017)

\bibitem{long}
{Shelhamer}, E., {Long}, J., {Darrell}, T.: Fully convolutional networks for
  semantic segmentation. IEEE Transactions on Pattern Analysis and Machine
  Intelligence  \textbf{39}(4),  640--651 (2017).
  \doi{10.1109/TPAMI.2016.2572683}

\bibitem{vgg19}
Simonyan, K., Zisserman, A.: Very deep convolutional networks for large-scale
  image recognition. arXiv 1409.1556  (09 2014)

\bibitem{Imbalance2019}
Smoliakov, D., Korotin, A., Erifeev, P., Papanov, A., Burnaev, E.:
  Meta-learning for resampling recommendation systems. In: Eleventh
  International Conference on Machine Vision (ICMV 2018); 110411S (2019). Proc.
  SPIE, vol. 11041 (2019). \doi{10.1117/12.2523103}

\bibitem{superpixels}
Sulimowicz, L., Ahmad, I., Aved, A.J.: Superpixel-enhanced pairwise conditional
  random field for semantic segmentation. 2018 25th IEEE International
  Conference on Image Processing (ICIP) pp. 271--275 (2018)

\bibitem{Szegedy2016Inceptionv4IA}
Szegedy, C., Ioffe, S., Vanhoucke, V.: Inception-v4, inception-resnet and the
  impact of residual connections on learning. In: AAAI. pp. 4278--4284 (2016)

\bibitem{ConformalMartingales2017}
Volkhonskiy, D., Burnaev, E., Nouretdinov, I., Gammerman, A., Vovk, V.:
  Inductive conformal martingales for change-point detection. In: Proc. of 6th
  Workshop COPA. PRML, vol.~60, pp. 132--153. PMLR (2017)

\end{thebibliography}

\end{document}